\documentclass[journal]{IEEEtai}
\usepackage[colorlinks,urlcolor=blue,linkcolor=blue,citecolor=blue]{hyperref}
\usepackage{array}
\usepackage{bibentry}  
\usepackage{xcolor,soul,framed} 

\colorlet{shadecolor}{yellow}
\usepackage[pdftex]{graphicx}
\graphicspath{{../pdf/}{../jpeg/}}
\DeclareGraphicsExtensions{.pdf,.jpeg,.png}

\usepackage{makecell}
\usepackage{graphicx}
\usepackage{subfigure}
\usepackage[normalem]{ulem}
\usepackage{url}
\usepackage{enumerate}
\usepackage{booktabs}
\usepackage{multirow}
\usepackage{xspace}

\usepackage{amsmath}
\usepackage{color}
\usepackage{threeparttable}
\usepackage{ulem}
\usepackage{amssymb}

\usepackage{float}
\usepackage[ruled,vlined,linesnumbered]{algorithm2e} 
\usepackage{amsmath} 
\usepackage{setspace}
\usepackage{caption}
\usepackage{subfigure}
\usepackage{fancyhdr} 
\usepackage{subcaption}

\def\cal#1{\mathcal{#1}}

\newcommand{\eg}{\textit{e.g.,}\xspace}

\newcommand{\encoder}{LHEncoder\xspace}
\newcommand{\retr}{HRetriever\xspace}
\newcommand{\agg}{TAggregator\xspace}

\newcommand{\Name}{LMHR\xspace}

\makeatother
\IEEEaftertitletext{\vspace{-3\baselineskip}}

\newcommand{\copyrightnotice}{%
  \scriptsize
  © 2025 IEEE. Personal use of this material is permitted.\newline
  Accepted for publication in \textit{IEEE Transactions on Artificial Intelligence}.\newline
  DOI: 10.1109/TAI.2025.3570676%
}

\fancypagestyle{firstpage}{
  \fancyhf{}                              
  \fancyfoot[C]{\copyrightnotice}          
}

\begin{document}
    \title{Leveraging  Long-Term Multivariate History Representation 
for Time Series Forecasting}
      \author{Huiliang Zhang,
      Di Wu,
           Arnaud Zinflou,~\IEEEmembership{Senior Member,~IEEE}, Stephane Dellacherie, \\Mouhamadou Makhtar Dione, and Benoit Boulet,~\IEEEmembership{Senior Member,~IEEE},
    
      \thanks{
    This work was supported in part by MITACS under Application
IT28431 and in part by Quebec’s Fonds de recherche Nature et technologies.
(\textit{Corresponding authors: Huiliang Zhang; Benoit Boulet}.)

      Huiliang Zhang, Di Wu and Benoit Boulet are with the Department of Electrical and Computer Engineering, McGill University, Montreal, QC H3A 0G4, Canada. (e-mail: huiliang.zhang2@mail.mcgill.ca; \{di.wu5,
    benoit.boulet\}@mcgill.ca;) Arnaud Zinflou, Stephane Dellacherie and Mouhamadou Makhtar Dione are with Hydro-Québec Research Institute, Montreal, QC H2Z 1A4, Canada. (e-mail: \{zinflou.arnaud, dellacherie.stephane, dione.mouhamadoumakhtar\}@hydroquebec.com)}} 

\markboth{IEEE Transactions on Artificial Intelligence, Vol. XX, No. X, Month 2025
}{H. Zhang \MakeLowercase{\textit{et al.}}: 
Leveraging Multivariate Long-Term History Representation 
for Time Series Forecasting}

\maketitle
\thispagestyle{firstpage}

\begin{abstract}
Multivariate Time Series (MTS) forecasting has a wide range of applications in both industry and academia. Recent advances in Spatial-Temporal Graph Neural Network (STGNN) have achieved great progress in modelling spatial-temporal correlations.
Limited by computational complexity, most STGNNs for MTS forecasting focus primarily on short-term and local spatial-temporal dependencies. 
Although some recent methods attempt to incorporate univariate history into modeling, they still overlook crucial long-term spatial-temporal similarities and correlations across MTS, which are essential for accurate forecasting.
To fill this gap,  we propose a framework called the \textbf{L}ong-term \textbf{M}ultivariate \textbf{H}istory \textbf{R}epresentation (\Name) Enhanced STGNN for MTS forecasting.
Specifically, a Long-term History Encoder (\encoder) is adopted to effectively encode the long-term history into segment-level contextual representations and reduce point-level noise. A  non-parametric Hierarchical Representation Retriever (\retr) is designed to include the spatial information in the long-term spatial-temporal dependency modelling and pick out the most valuable representations with no additional training. A Transformer-based Aggregator (\agg) selectively fuses the sparsely retrieved contextual representations based on the ranking positional embedding efficiently. 
Experimental results demonstrate that \Name outperforms typical STGNNs by 10.72\% on
the average prediction horizons and state-of-the-art methods by 4.12\% on several real-world datasets. Additionally, it consistently improves prediction accuracy by 9.8\% on the top 10\% of rapidly changing patterns across the datasets.

\end{abstract}

\begin{IEEEImpStatement}
Multivariate time series (MTS) data is ubiquitous in our daily lives, and accurate forecasting is highly valuable for decision-making. However, recent advances in Spatial-Temporal Graph Neural Networks (STGNNs) are limited by computational complexity, primarily focusing on local spatial-temporal dependencies while neglecting crucial long-term spatial-temporal similarities and correlations across MTS. The simple yet effective long-term multivariate history representation-enhanced STGNN framework introduced in this paper overcomes these limitations. Extensive experiments on several real-world datasets demonstrate significant improvements in forecasting performance, along with explainable results. The proposed general framework also shows promising application prospects and paves the way for efficiently improving existing STGNN models with minimal changes.
\end{IEEEImpStatement}
\section{Introduction}
\IEEEPARstart{A}{ccurate} forecasting of Multivariate Time Series (MTS) is incredibly valuable for improving decision-making, attracting broad interest across various academic and industrial fields.
For instance, accurate and reliable forecasts of traffic flow and speed can guide real-time route planning and help avoid congestion~\cite{wang2020traffic}. Similarly, accurate predictions of electricity consumption are crucial for managing peak demand periods and minimizing economic losses~\cite{deb2017review}.
With the development of deep learning, many models have been proposed and achieved superior performance in MTS forecasting~\cite{MTS_survey,Onesize_INNLS}.
Among them, the most recent efforts formalize MTS into spatial-temporal graph data ~\cite{GWNet,spatial-temporal-aware,sp_encoder_traffic,dmg}, acknowledging the intricate temporal dynamics and spatial relationships that are mutually influenced between nodes.

\begin{figure}[hbp!]
 \includegraphics[width=\linewidth]{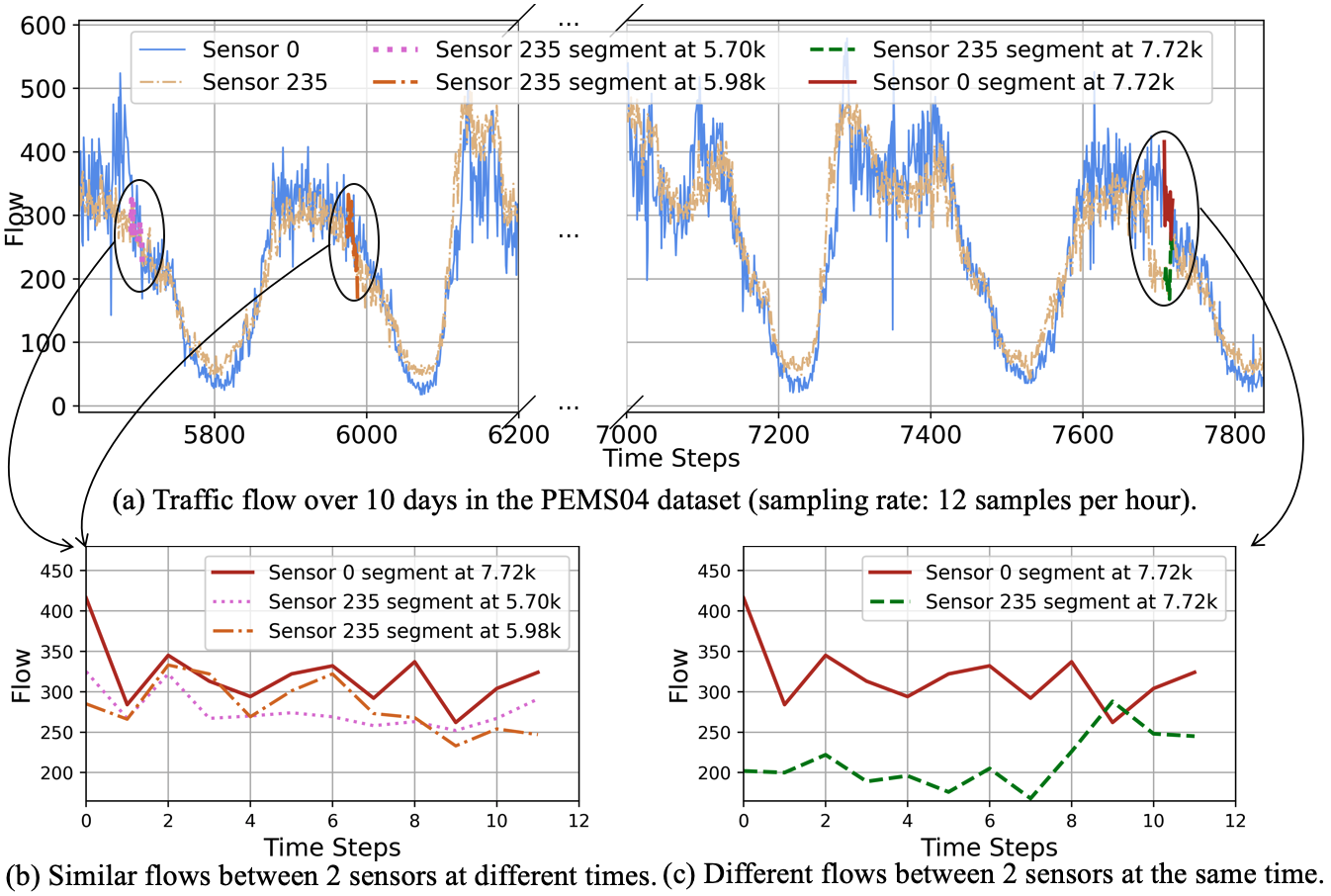}
 \caption{Traffic flow examples from PEMS04 dataset. (a) Two sensors (0 and 235) show complex and various spatial-temporal correlations and similarities over a long-term history. (b) Similar temporal patterns from sensors 0 and 235 at different times, showing potential value from  long-term multivariate history. (c) Different temporal patterns (magnitude and trend) from sensors 0 and 235 in a short-term window, demonstrating the unreliability of current methods that rely solely on short-term multivariate history.}
 \label{example}
\end{figure}

Recently, Spatial-Temporal Graph Neural Networks (STGNNs) have become prevalent as MTS forecasting methods 
due to their cutting-edge performance \cite{stgcnyu2017spatio,heriti_SPGNN_traffic}. 
In addition, emerging research in this field aims to construct more precise graph structure~\cite{dmg,stemgnncao2020spectral,shang2021discrete, ye2022learning} and fuse spatial-temporal data with elaborate graph neural network 
\cite{sp_encoder_traffic,stgcnyu2017spatio,cluster-traffic-flow,wu2020connecting,d2stgnn,dgcrnli2021dynamic}.
Nevertheless, due to the large computational complexity introduced by the spatial-temporal graph architecture, these works mainly focus on capturing spatial-temporal dependencies from short-term input  (e.g., data over the past hour). 
As a result, they tend to learn local prominent relevance and correlations, short-term spatial-temporal patterns and similarities \cite{heriti_SPGNN_traffic,chen2023multi_scale,Multiview-Time-Series-ITAI}.

However, the intricate and complex patterns that emerge and are influenced over an extended historical context are also crucial for accurate forecasting.
Some recent work has attempted to convert long-term univariate time series into segment embeddings using the Transformer architecture~\cite{transformervaswani2017attention, shao2022pre, patchTst, sparse-transformer-ITAI}. However, the empirical improvements in forecasting capabilities are modest, as these models primarily capture temporal dependencies but neglect spatial factors over the long term. 
This oversight of spatial-temporal similarities and relations, with limited ability to extract spatial-temporal representations, hinder the models' ability to efficiently mine and utilize relevant patterns and trends that occurred historically, especially those rare and rapidly changing ones.

To illustrate, we take a traffic flow system as an example, where each sensor is treated as a series variable. 
Fig. \ref{example}(a) depicts traffic flows over 10 days from two adjacent sensors: 0 (blue) and 235 (orange).
We can observe obvious correlations and similarities between sensor 0 and sensor 235 from the long-term history. Furthermore, the data from sensor 0 at the 7.72k time step shows a similar pattern (magnitude and trend) compared with a segment from sensor 235 at 5.70k and 5.98k time steps, as shown in Fig. \ref{example}(b). On the contrary, sensor 0 appears considerably different from sensor 235 in the same time window at the 7.72k time step as shown in Fig. \ref{example}(c).  This implies that predictions based solely on short-term multivariate history, as seen in previous works, may be unreliable \cite{GWNet,stemgnncao2020spectral,d2stgnn,shao2022pre,patchTst,knnmts}. Furthermore, similar but sporadic trends, or  even rare and rapidly changing patterns are also more likely to be detected from long-term contexts, even if they are sparsely distributed. Moreover, temporal dependencies and dynamics are deeply influenced by spatial factors such as geographical adjacency, road connections, and functional similarities. These complexities are far greater than what is observed from short-term time series. Thus, to achieve accurate forecasting, it is essential to consider both temporal and spatial relations, incorporating useful patterns from the long-term multivariate history.


Nevertheless, there are several challenges to directly absorb long-term multivariate history 
and consider the complex spatial-temporal dependencies and relations 
in MTS forecasting models. First, most current STGNNs
are limited to the short-term input data (\eg past 1 hour), since their computational complexity usually increase linearly or even quadratically with the length of the input \cite{stgcnyu2017spatio,dgcrnli2021dynamic}, and the computational burden is further aggravated by the number of variables \cite{MTS_survey}.
Second, long-term multivariate history brings redundant information in both spatial and temporal dimensions which will introduce noisy input to the model \cite{transformervaswani2017attention,shao2022pre, sparse-transformer-ITAI}. As shown in Fig. \ref{example}(a), we can observe a large number of dissimilar historical segments in the long-term data from sensor 235, compared with the segment from sensor 0 at time step 7.72k. Even with the help of segment-level transformer, it would be unaffordable to analyze every individual attention weight for all MTS segments in large datasets and will also bring lots of noisy inputs.
In summary, how to efficiently identify and leverage spatial-temporal dependencies and relations from long-term multivariate histories in MTS forecasting remains an unresolved challenge.


To fill the gap and address the above challenges, we propose a simple yet effective framework which strengthens STGNNs with spatial-temporal information and relations over long-term multivariate history for MTS forecasting, with the name called \textbf{L}ong-term \textbf{M}ultivariate \textbf{H}istory \textbf{R}epresentation (\Name) Enhanced Spatial-temporal Graph Neural Network.
The \Name framework is composed with three modules: a \textbf{L}ong-term \textbf{H}istory \textbf{Encoder} (\encoder), a \textbf{H}ierarchical Representation  \textbf{Retriever} (\retr), and a \textbf{T}ransformer-based  \textbf{Aggregator} (\agg). For the first challenge, multivariate MTS series are separately encoded by the \encoder to obtain contextual segment-level representations. Dense segment-level representation incorporates sub-series points together to reduce the input length and reduce point-level noise. 
For the second challenge, the \retr 
hierarchically searches potential useful but sparsely distributed information from series-level and segment-level using the contextual-aware representations 
encoded by \encoder.  In this way, the spatial information is considered into
the correlations embedding.
Also, \retr precisely identifies the top-$k$ similar representations from all multivariate MTS segments without additional training, reducing noise from the  long-term multivariate input, which contains valuable information but also significant redundancy. The retrieved segments' contextual representations, combined with their corresponding ranking position embeddings, are then fused by \agg to efficiently utilize the useful information.
\Name also leverages the segment-level representation 
to generate an adjacency matrix, which is used to guide the Graph Structure Learning (GSL) in case the pre-defined graph in STGNN is incomplete. 
\begin{table}[htbp!]
\centering
\small 
\caption{Variables and Descriptions}
\label{tab:variables}
\begin{tabular}{p{0.072\textwidth}|p{0.37\textwidth}} 
\toprule
\textbf{Variable} & \textbf{Description} \\ 
\midrule
$T$ & Number of time steps. \\ 
$N$ & Number of nodes (series).\\ 
$C$ & Number of data channels for each node. \\ 
$\mathcal{X}$ & Multivariate time series tensor, $\mathcal{X} \in \mathbb{R}^{T \times N \times C}$. \\ 
$\mathcal{G}$ & Graph network with nodes $V$ and edges $E$. \\ 
$\mathbf{A}$ & Graph adjacency matrix, $\mathbf{A} \in \mathbb{R}^{N \times N}$. \\ 
$T_h$ & Number of historical time steps. \\ 
$T_f$ & Number of future time steps to predict. \\ 
$\mathcal{Y} $ & Future time series tensor, $\mathcal{Y} \in \mathbb{R}^{T_f \times N \times C}$. \\ 
$L$ & Length of the long-term history for a node. \\ 
$L_s$ & Length of a segment in long-term history. \\ 
$P$ & Number of segments in long-term history. \\ 
$l$ & Length of the non-overlapping region between consecutive segments. \\ 
$\mathbf{E}_j^i$ & Temporal embedding of $j$-th segment of $i$-th series. \\ 
$d$ & Hidden state dimension. \\ 
$\mathbf{H}^i$ & Context-aware segment-level representation of $i$-th series. \\ 
$K_n$ & Number of similar series retrieved at series-level. \\ 
$K_s$ & Number of similar segments retrieved at segment-level. \\ 
$\mathbf{H}^i_P$ & Representation of the last segment of $i$-th series. \\ 
$\mathbf{A}^r$ & Adjacency matrix constructed from series-level retrieval, $\mathbf{A}^r \in \mathbb{R}^{N \times N}$. \\ 
$\mathbf{R}_k$ & Ranking position embedding for retrieved segment. \\ 
$\mathbf{AO}_{P+1}^i$ & Output of the aggregator for $(P+1)$-th segment of  $i$-th series. \\ 
$\mathbf{H}^i_{stgnn}$ & Short-term spatial-temporal hidden state of $i$-th node. \\ 
$\mathbf{H}^i_P$ & Long-term temporal contextual representation of $i$-th node. \\ 
$\mathbf{H}_{final}$ & Final fused hidden state for forecasting. \\ 
$\mathbf{\Theta}$ & Parameter matrix of the Bernoulli distribution. \\
\bottomrule
\end{tabular}
\end{table}
Notably, \Name is a general framework that can absorb and enhance almost arbitrary
STGNNs for MTS forecasting. In summary, the main contributions are the following:
\begin{itemize}
    \item We recognize the importance of utilizing long-term multivariate history in MTS forecasting. We propose a simple yet effective \Name framework to enhance current STGNNs by including the spatial-temporal information into the correlations embedding 
    over the long-term multivariate history, leveraging the representations from \encoder. 

    \item We design a non-parametric \retr which efficiently extracts useful but sparsely distributed information from both series and segment levels
    with no additional training, and reduces the noise input from the long-term multivariate history.
    Plus, a \agg is designed to better pick out and selectively merge valuable retrieved representations. 
    
    \item We conduct extensive evaluations on several real-world datasets, demonstrating the interpretability of \Name with detailed illustrations.
 Experimental results demonstrate that our proposed framework outperforms typical STGNNs by 10.72\% on
the average prediction horizons and state-of-the-art methods by 4.12\%. Additionally, it improves prediction accuracy by 9.8\% on the top 10\% of rapidly changing patterns across the datasets.

\end{itemize}
\begin{figure*}[htbp!]
 \centering
 \setlength{\abovecaptionskip}{-0.1cm}
 \setlength{\belowcaptionskip}{-0.4cm}
 \includegraphics[width=0.95\linewidth]{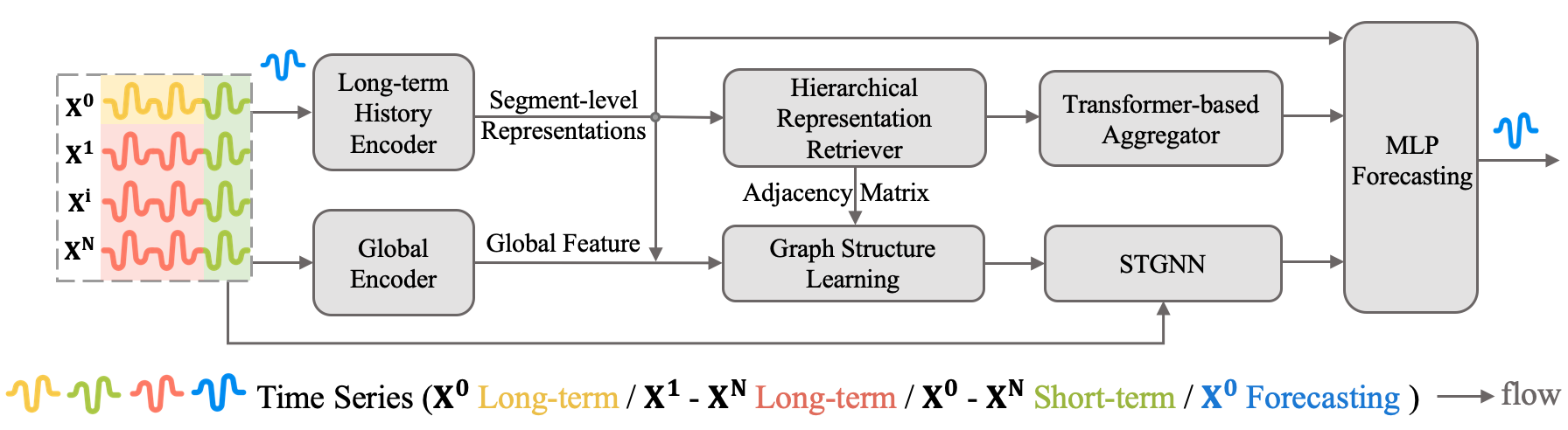}
 \caption{
 {\color{black}
 The overview of the proposed \Name framework for MTS forecasting.
The yellow, red, green, and blue curves represent long-term time series information from the node itself, from other nodes, short-term information from MTS, and forecasting-related information, respectively.
   Long-term histories of multivariate MTS series are separately encoded by the \encoder.
    \retr 
hierarchically searches potential useful information and outputs
the top-$k$ similar representations from all nodes to \agg. \agg fuses the retrieved segments' contextual representations with corresponding ranking position embeddings.
\Name also leverages the representations from \retr
to generate an adjacency matrix used in STGNN to extract the short-term dependencies. The representations from \encoder, \retr and \agg are fused together in final forecasting.
 }
 }
 \label{fig:framework}
\end{figure*}

\section{Methodology}

\subsection{Problem Statement}
Multivariate time series can be denoted as a tensor $\mathcal{X}\in\mathbb{R}^{T\times N\times C}$, where $T$ is the number of time steps, $N$ is the number of nodes (series), and $C$ is the number of data channels for each node. The graph network is described as $\mathcal{G}=(V, E)$, where $V$ is the set of $|V|=N$ nodes, $E$ is the set of $|E|=M$ edges indicating the connectivity of nodes.
The graph adjacency matrix derived from the  network is denoted by $\mathbf{A} \in \mathbb{R}^{N\times N}$. 
Given $N$
time series with $T_h$ historical signals $\mathcal{X}\in\mathbb{R}^{T_h\times N\times C}$, time series forecasting aims to predict next $T_f$ future values $\mathcal{Y} \in \mathbb{R}^{T_f \times N \times C}$, based on the network $\mathcal{G}$ and history signals $\mathcal{X}$. The description of frequently used variables in this paper can be found in Table \ref{tab:variables}.

\subsection{Overview of Framework}
The overview of \Name framework is shown in Fig. \ref{fig:framework}, where long-term MTS are used for forecasting at each node. 
The long-term history from $i$-th node is represented as $\mathbf{X}^{i} \in \mathbb{R}^{L \times C}$ and we take the 0-th node $\mathbf{X}^0$ as a forecasting target to illustrate the workflow of \Name. As illustrated in the middle-top of Fig. \ref{fig:framework}, raw time series is first transformed by the Long-term History Encoder (\encoder) to learn the context-aware segment-level representation and long-term temporal dependencies.
To efficiently consider spatial information in the dependency relations embedding,
the  non-parametric Hierarchical Representation Retriever (\retr) first retrieves the top $K_n$ relevant series
over $\mathbf{X}^1$ to $\mathbf{X}^N$ from series-level and then gets top
$K_s$  segment-level representations. 
\retr also provides a semi-supervised adjacency matrix for Graph Structure Learning (GSL) and segment similarities for the Transformer-based  Aggregator (\agg) to capture the spatial-temporal relations over long-term history. Then \agg generates a fused representation based on the top $K_s$ similarities, combining their ranking position embeddings, $\mathbf{X}^0$ long-term segment representations, and the forecasting position embedding. In the middle-bottom of Fig. \ref{fig:framework}, a global encoder provides a stable global feature corresponding with dynamic segment representation for GSL in case the pre-defined graph is not complete. The learned graph structure will be sent to STGNN to generate a spatial-temporal graph representation and extract short-term spatial-temporal dependencies. 
\Name is compatible with almost any STGNN model and we illustrate it mainly based on the classic Graph WaveNet in this section.
\subsection{Long-term History Encoder}
One of the main objectives of MTS forecasting is understanding the correlation between data at different timesteps. However, unlike words in a sentence, an individual timestep lacks inherent semantic meaning. Comparable time series patterns often emerge within similar time spans due to analogous time series conditions. Therefore, extracting local semantic information from time series is crucial for analyzing their interconnections. Furthermore, point-wise short-term input may not adequately capture all potential time series trends, such as cyclical patterns observed a few days earlier.

Given that long-term time series contain more temporal patterns and correlations, 
directly encoding thousands of points in neural networks can pose challenges due to its significant resource demands. As a result, the \encoder first divides the raw series $\mathbf{X}^{i}$ into segments and then transforms them into temporal embeddings. This provides the model with a look-back window over its long-term history.
In \encoder, we design a soft-break which allows upcoming segmentations to overlap with previous ones, and the $\mathbf{X}^{i}$ is divided into $P$ segments $\mathbf{S}^i = [\mathbf{S}_0^i, \mathbf{S}_1^i, ..., \mathbf{S}_P^i]$. Each segment $\mathbf{S}_j^i \in \mathbb{R}^{L_s \times C}$ contains $L_s$ time steps, and the number of segment $P$ could be calculated as follows:
\begin{equation}
    P = \lfloor \frac{L - L_s}{l}  \rfloor +2,
\end{equation}
where $l$ is the length of non-overlapping region between two continues segments. Here, we pad $l$ repeated numbers of the first value $\mathbf{S}_0^i$ at the beginning of the original sequence before we separate the series into segments. With the retain of $l$ timestep overlap in a soft break way, the useful temporal patterns are preserved to the fullest extent.
 The segments could also be non-overlapped by setting $l=L_s$, which is a special case of the proposed segmentation. To simplify the explanation of our methodology, we assume $C=1$ in the illustrations.

Then the temporal embedding of $j$-th segment of $i$ series $\mathbf{E}^i_j$ is generated by a linear layer with learnable positional embedding $\mathbf{p}_j$:
\begin{equation}
    \mathbf{E}_j^i = \mathbf{W}\cdot\mathbf{S}^i_j + \mathbf{b} + \mathbf{p}_j,
\end{equation}
where $\mathbf{W}\in\mathbb{R}^{d\times L_s }$ and $\mathbf{b}\in\mathbb{R}^d$ are learnable parameters, $d$ is the hidden state dimension. 

Since the segment embeddings are blind to the context information beyond its data window of $L_s$, it might be difficult for the model to distinguish valuable segments in different contexts and learn long-term temporal dependencies. So we utilize the multiple temporal Transformer layers to give $j$-th segment of series $i$ rich contextual information $\mathbf{H}^i_j$. Each temporal Transformer layer has a multi-head self-attention network (MSA) and a position-wise fully connected feed-forward network (FFN). A residual connection around layers with layer normalization is also added:
\begin{equation}
 \begin{aligned}
 & \mathbf{U}^i=\mathrm{LayerNorm}(\mathbf{E}^i + \mathrm{MSA}(\mathbf{E}^i)), \\
& \mathbf{H}^i=\mathrm{LayerNorm}(\mathbf{U}^i + \mathrm{FFN}(\mathbf{U}^i)),
\end{aligned}
 \label{eq:transformer}
\end{equation}
where $\mathbf{E}^i=(\mathbf{E}^i_1,\mathbf{E}^i_2,...\mathbf{E}^i_{P-1},\mathbf{E}^i_{P})$ are the input temporal embeddings and 
we use four stacked Transformer layers to get the final contextual representations.

Moreover, the strategy of  segmenting the time series series into soft-break segments 
serves multiple purposes.
First, segments are more stable and contain better contextual information within their window length than separate points.
Second, it significantly reduces the length of the sequence input length for the temporal Transformer layer, which makes the long-term history encoding process more efficient.
Third, the segment-level representations from the long-term history encoder could be easily merged with the STGNN models, since they usually take the short-term history as input (past twelve points), which could be seen as the last segment.

\subsection{Hierarchical Representation Retriever}
\begin{figure}[tbp!]
 \centering
 \setlength{\belowcaptionskip}{-0.7cm}
 \includegraphics[width=0.48\textwidth]{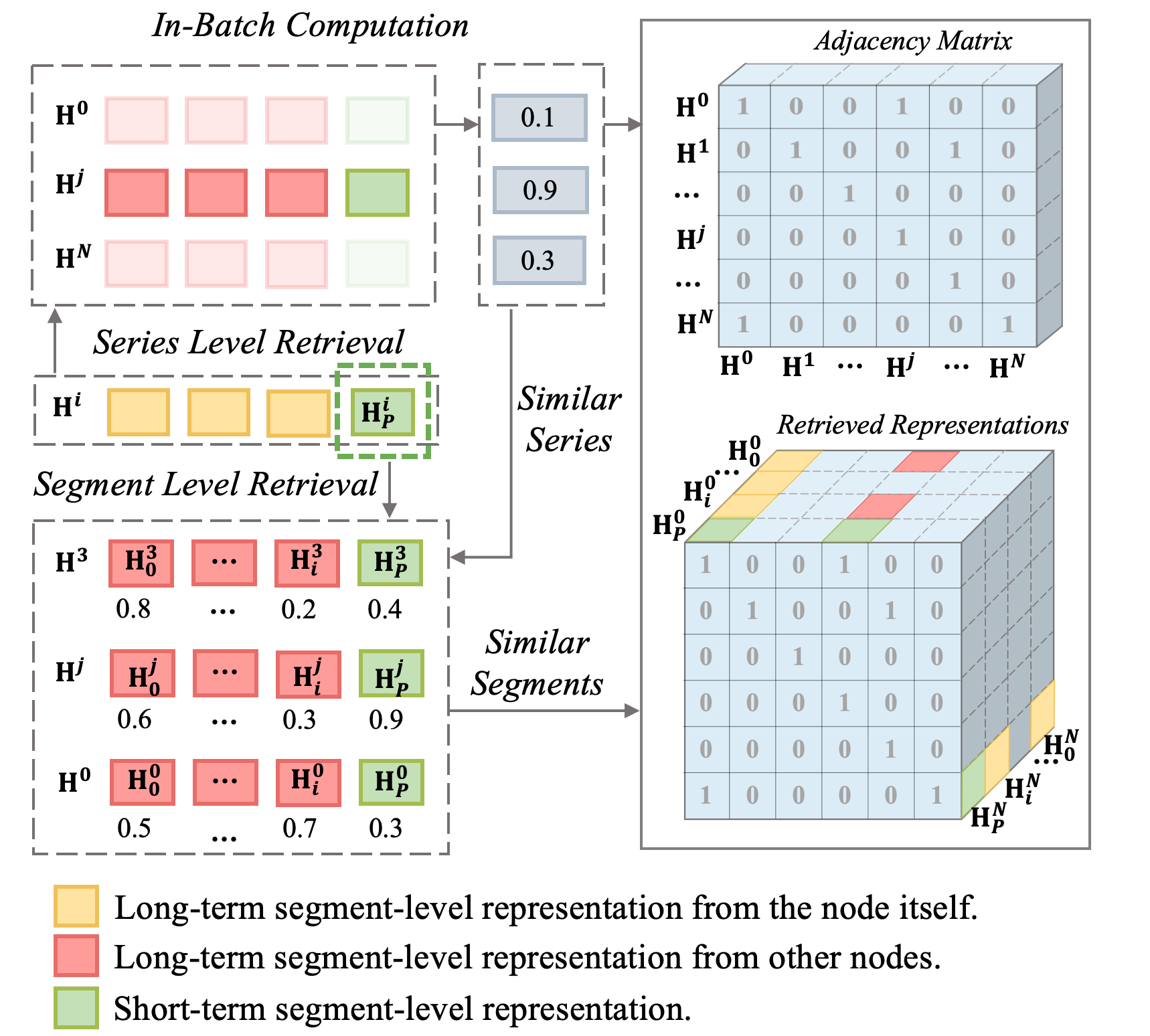}
 \caption{
 {\color{black}
 The Hierarchical Representation Retriever.
  The colors yellow, red and green represent long-term information from the node itself, from other nodes, and the short-term segment-level  representation nearest the prediction, respectively.
  \retr first finds top $K_n$ similar series through series-level retrieval and constructs the adjacency matrix, then it picks out the top $K_s$ similar segment representations from the retrieved similar series using segment-level retrieval.
  } } \label{fig:retriever}
\end{figure}
Directly absorbing long-term multivariate history and considering the complex
spatial-temporal dependencies and relations in MTS forecasting models is challenging.
Besides, the redundant information and noise input from the long-term MTS data could also lead to inefficient learning. 
We hypothesize that time series series that are closer in representation space are
more likely to be followed by the same future series.
Hence, we propose the \retr to hierarchically search potential useful information from series-level and segment-level using the contextual-aware representations 
encoded by \encoder. The spatial information is also considered in
the correlations embedding efficiently in a non-parametric way. 

As shown in the top left of Fig. \ref{fig:retriever}, the \retr first uses the long-term representations $\mathbf{H}^i$  from each nodes (yellow block)
to find their top $K_n$ similar representations from series-level (red block). 
The similarity could be calculated in batch using cosine similarity 
\begin{equation}
\label{eq:batch_cos_similrity}
    \text{{cosine\_similarity}}(\mathbf{H}, \mathbf{H}^T) = \frac{\mathbf{H} \cdot \mathbf{H}^T}{\|\mathbf{H}\| \|\mathbf{H}^T\|},
\end{equation}
and ranked to get the top $K_n$ similar series.
This process would also produce pair-wise similarities between all variables and then be converted into a $N \times N$ adjacency matrix $\mathbf{A}^r \in \mathbb{R}^{N \times N}$. 
For example, if we take $K_n=1$, and the series-level similarity calculated from sensor 3 is the top 1 among others for sensor 1, then we recognize those two series are similar and sensors are adjacent. Furthermore, the two sensors are linked in the adjacency matrix, as shown in the top right of Fig. \ref{fig:retriever}. 

Next, the segment-level representation nearest the prediction (green block) is leveraged to pick out the top $K_s$ valuable segment representations from the retrieved similar series. 
We calculate the similarity in-batch between last-segment representations $\mathbf{H}_P^i$ of the target node and the retrieved similar series' segment-level representation $\mathbf{H}^i_j$, respectively. Then we rank these similarities and select the top $K_s$ similar segments' representations.
As illustrated in the bottom of Fig. \ref{fig:retriever}, we use the $\mathbf{H}_P^0$ from sensor 0 to find similar segment representations in $\mathbf{H}^3$ from sensor 3. The cubic version of the adjacent matrix shows the segment-level retrieval results of $N$ variables, which could be computed simultaneously. 
Notably, the \retr does not introduce any trainable parameters and is computation efficiently by processing the long-term MTS hierarchically, and the spatial-biased temporal dependency is extracted explicitly.
The pseudocode of the \retr algorithm is outlined in Appx. 
\ref{appendix_algo}.

\subsection{Transformer-based Aggregator}
\begin{figure}[tbp!]
  \centering
  \setlength{\belowcaptionskip}{-0.5cm}
  \includegraphics[width=0.45\textwidth]{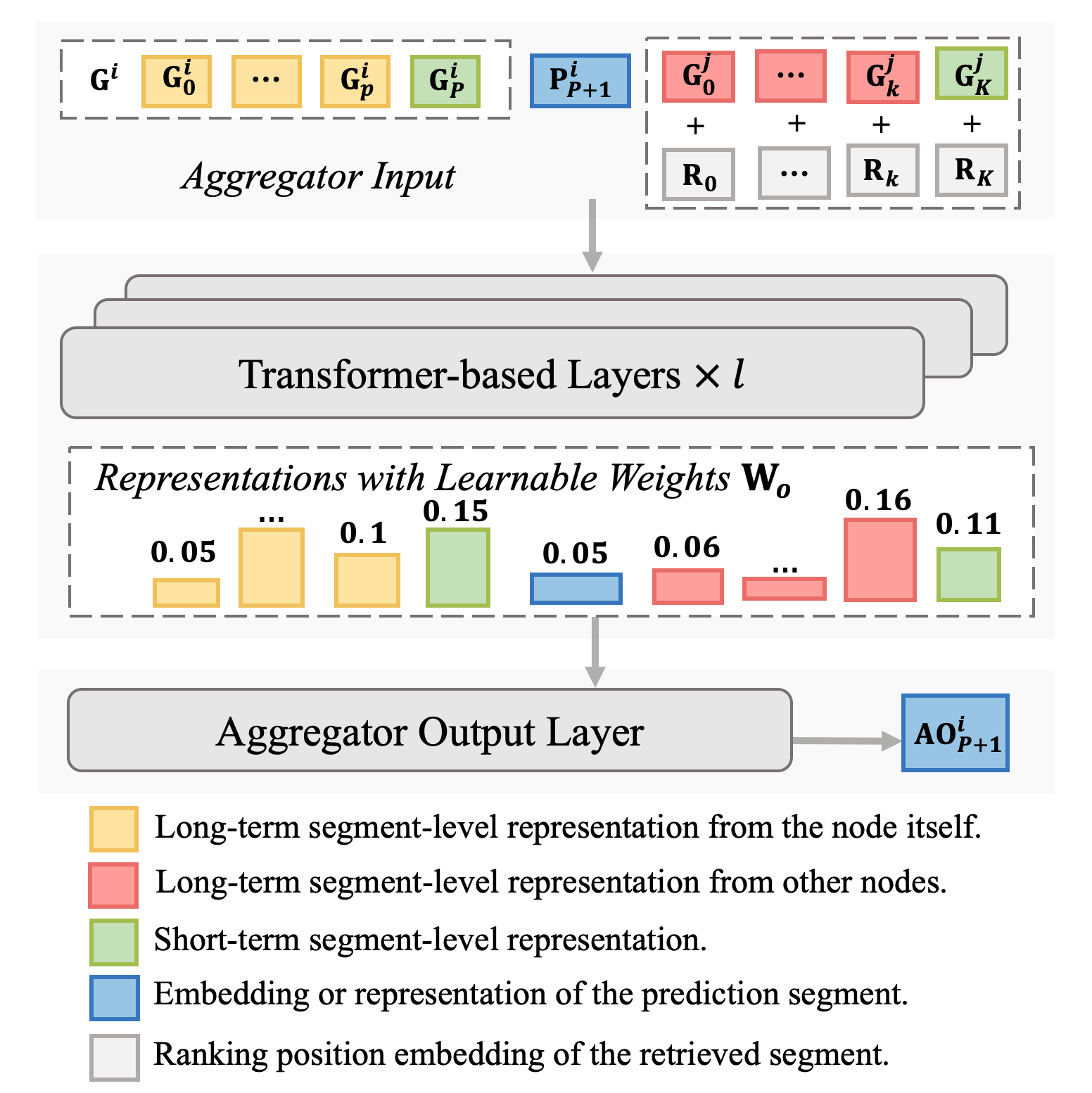}
  \caption{
  {\color{black}
The Transformer-based Aggregator. The colors yellow, red and green in Fig. \ref{fig:aggregator} have the same meaning as those in Fig. \ref{fig:retriever}. The blue represents the embedding or representation of the prediction segment, and the grey is the ranking position embedding of the retrieved segment.
  The \agg utilizes positional embedding of the target segment to absorb retrieved segment representations from MTS long-term history using the Transformer method.
  Learnable weights are dynamically updated during training.}
  }
  \label{fig:aggregator}
\end{figure}
After the \retr finds the top $K_s$ similar segment representations efficiently, the \agg is proposed to absorb them based on the positional embedding of the target segment and retrieved similar historical segment representations. 
The \agg is designed to capture the sequential trend between future and historical data by adding a learnable positional embedding to the \((P+1)\)-th forecasting segment, as illustrated in Fig. \ref{fig:aggregator}. 
For the $i$-th time series, the \agg first projects the target node's segment representations $\mathbf{H}^i$ into $\mathbf{G}^i = [\mathbf{G}^i_0, \mathbf{G}^i_1, ..., \mathbf{G}^i_P]$ with a linear layer:
\begin{equation}
    \mathbf{G}^i_j = \mathbf{W}_g\mathbf{H}^i_j + \mathbf{b}_g,
\end{equation}
where  $ \mathbf{W}_g \in \mathbb{R}^{d \times d}$, $\mathbf{H}^i_j\in \mathbb{R}^{d \times 1}$, and $\mathbf{b}_g \in \mathbb{R}^{d}$. 

Then the \agg concatenates the historical segment representations $\mathbf{G}^i$, the target position embedding $\mathbf{p}^i_{P+1}$ and the retrieved segment representations combined with their ranking position embeddings:
\begin{equation}
    \mathbf{AI}^i = \mathbf{G}^i_0 || \mathbf{G}^i_1 || ... || \mathbf{G}^i_P || \mathbf{p}^i_{P+1} || (\mathbf{G}^j_k + \mathbf{R}_k)||...||( \mathbf{G}^j_K + \mathbf{R}_K),
\end{equation}
where $\mathbf{AI}^i$ is the aggregator input of the $i$-th series, $\mathbf{R}_k$ is the ranking position embedding of the $k$-th retrieved segment, and $\mathbf{p}^i_{P+1}$ is the target $(P+1)$-th position embedding. The sequence position embedding is different from the ranking position embedding, since the former is derived from the continuous segment position and the latter is derived from the \retr.

The $\mathbf{AI}^i$ is fed into $l$  Transformer layers. The \agg projects the $(P+1)$-th hidden state after Transformer into the output $\mathbf{AO}^i_{P+1}$ with a linear layer:
\begin{equation}
\mathbf{AO}^i_{P+1} = \mathbf{W}_o transformer(\mathbf{AI}^i)+\mathbf{b}_o,
\end{equation}
where the $transformer$ is the same with Eqn. \ref{eq:transformer}, 
$\mathbf{W}_o $ and $\mathbf{b}_o$ are both learnable parameters.
The output $\mathbf{AO}_{P+1}$ of \agg selectively picks out important representations from the long-term histories of each node and retrieved segment representations from other nodes, which also effectively reduces random noise through weighted averaging.


\subsection{Forecasting}
In forecasting, the \encoder, non-parameter \retr, and \agg are utilized to extract useful information from
long-term MTS. Furthermore, the STGNN is also used in \Name to extract short-term spatial-temporal dependencies to further improve forecasting accuracy.
As shown in Fig. \ref{fig:framework}, for $i$-th node, \Name leverages the short-term spatial-temporal hidden states $\mathbf{H}^i_{stgnn} $
from STGNNs, the long-term temporal contextual representation $\mathbf{H}^i_P $
from \encoder, the long-term spatial-temporal representation $\mathbf{AO}^i_{P+1} $
from \agg to generate the fused hidden states $\mathbf{H}_{final}^i$
for forecasting: 
\begin{equation}
\begin{aligned}
& \mathbf{H}_{final}^i= MLP(\mathbf{H}^i_{stgnn}) + MLP(\mathbf{H}^i_P) + MLP(\mathbf{AO}^i_{P+1}), \\\mathbf{
}& \hat{\cal{{Y}}^i} = MLP(\mathbf{H}^i_{final}),
\end{aligned}
\end{equation}
where the $\hat{\cal{Y}^i} \in \mathbb{R}^{T_f}$ is the $T_f$ future points of $i$-th time series and $MLP$ is a Multi-Layer Perceptron. 

Then \Name computes the loss function between predictions and real values for all time series of $T_f$ steps and $C$ channels:
\begin{equation}
    \cal{L}_{regression} = \cal{L}(\hat{\cal{Y}}, \cal{Y})=\frac{1}{T_fNC}
    \sum_{j=1}^{T_f}\sum_{i=1}^{N}\sum_{k=1}^{C}|\hat{\cal{Y}}^i_{ik} - \cal{Y}^i_{jk}|, 
    \label{loss}
\end{equation}

\Name is a general framework that can absorb
and enhance almost arbitrary STGNNs.
As our backend, we have chosen a representative method, namely
Graph WaveNet \cite{GWNet}, which combines graph convolution with dilated causal convolution to capture spatial-temporal dependencies.
We refer to the GraphWaveNet paper \cite{GWNet} for more details to get $\mathbf{H}^i_{stgnn}$ based on the last history segment $\mathbf{S}^i_P$.

Considering that most STGNNs need a graph structure to capture the spatial and temporal relationships, we introduce Graph Structure Learning (GSL)  to guide the adjacency matrix learning in case the predefined graph is missing or incomplete. 
The GSL aims to learn a discrete sparse graph, where $\mathbf{\Theta}_{ij}$ parameterizes the Bernoulli
distribution from which the discrete dependency graph $\mathbf{A}$ is sampled:
\begin{equation}
    \begin{aligned}
        & \mathbf{\Theta}_{ij} = MLP(\mathbf{Z}^i, \mathbf{Z}^j), \\
        & \mathbf{Z}^i = MLP(\mathbf{AO}^i_{P+1}) + \mathbf{F}^i_G, \\
        & \mathbf{F}^i_G = MLP(Conv(\mathbf{X}^i_{train})),
    \end{aligned}
\end{equation}
where $\mathbf{F}_G^i$ is the global feature from the Global Encoder shown in Fig. \ref{fig:framework}, which is a convolutional neural network. The Global Encoder takes all training data of $i$-th series $\mathbf{X}^i_{train}$ as input to produce a robust and stable learning process. $\mathbf{\Theta}_{ij}$ comes from mixed hidden state $\mathbf{Z}^i, \mathbf{Z}^j$ of $i$-th and $j$-th time series. The mixed hidden state combines the dynamic representation $\mathbf{AO}^i_{P+1}$ from \agg and the global feature $\mathbf{F}^i_G$ from Global Encoder. The \Name uses the calculated adjacency matrix $\mathbf{A}^r \in \mathbb{R}^{N \times N}$ from \retr to regularize the learning of $\mathbf{\Theta}$. Specifically, a cross-entropy loss between $\mathbf{\Theta}$ and $\mathbf{A}^r$ is added as an auxiliary loss along with the forecasting regression loss:
\begin{equation}
    \mathcal{L}_{graph} = \sum_{ij}-\mathbf{A^r}_{ij}\log\mathbf{\Theta}^{'}_{ij}-(1-\mathbf{A^r}_{ij})\log(1-\mathbf{\Theta}^{'}_{ij}),
\end{equation}
where $\mathbf{A}^r_{ij}$ is the pairwise relation between $i$-th and $j$-th time series, which is 0 for the unconnected or 1 for the connected and $\mathbf{\Theta}^{'}_{ij}=\text{softmax}(\mathbf{\Theta}_{ij})\in\mathbb{R}$ is the normalized probability.
To produce the $\mathbf{A}_{ij}$ sent into STGNNs, the \Name apply the Gumbel-Softmax reparametrization \cite{Gumbel1} trick to make the sampling operation from $\mathbf{\Theta}_{ij}$ trainable.
\begin{equation}
    \mathbf{A}_{ij}=\text{softmax}((\mathbf{\Theta}_{ij}+\mathbf{g})/\tau),
\end{equation}
where $\mathbf{g}\in \mathbb{R}^2$ is a vector of i.i.d. samples drawn from a $\text{Gumbel(0,1)}$ distribution \cite{Gumbel2}.
$\tau$ is the softmax temperature parameter.
Finally, the forecasting regression loss $\mathcal{L}_{regression}$ and the regularization graph loss $\mathcal{L}_{graph}$ are added with a hyper-parameter $\lambda$ to do an end-to-end training:
\begin{equation}
    \mathcal{L} = \mathcal{L}_{regression} + \lambda \mathcal{L}_{graph}.
    \label{full_loss}
\end{equation}

\section{Experiments}
In this section, we conduct experiments on several real-world datasets to illustrate the performance of the proposed \Name. Moreover, we present comprehensive experiments to analyze the ability, the impact of each module and different hyper-parameters in the \Name. 

\subsection{Experimental Setups}
\label{exp_set}
\textbf{Datasets}.
Following previous works~\cite{ GWNet,wu2020connecting, d2stgnn, shao2022pre, stidshao2022spatial}, we conduct experiments on five commonly used MTS datasets: PEMS-BAY, PEMS04, PEMS07, PEMS08 and Electricity.
The statistics of four datasets are summarized in Table \ref{tab:datasets} and traffic datasets come with a predefined graph.
For a fair comparison, we follow the dataset division in previous works \cite{shao2022pre, stidshao2022spatial}. For PEMS-BAY, we use 70\% of data for training, 10\% of data for validation, and the remaining 20\% for test~\cite{GWNet,2017DCRNN}.
For PEMS04, PEMS07, PEMS08 and Electricity, we use 60\% of data for training, 20\% of data for testing, and the remaining 20\% for validation~\cite{2021ASTGNN, 2019ASTGCN}.  
\begin{table*}[ht]
\centering
\caption{Statistics of datasets.}
\label{tab:datasets}
\small
\begin{tabular}{c|c|c|c|c|c}    
\toprule
\textbf{Dataset}&\textbf{Type} &\textbf{Samples} & \textbf{Node} & \textbf{Sample Rate} & \textbf{Time Span} \\
\midrule
{PEMS-BAY} & Traffic speed & 52116 & 325 &5mins & 6 months\\
{PEMS04}   & Traffic flow & 16992 & 307 &5mins & 2 months\\
{PEMS07}   & Traffic flow & 28224 & 883 &5mins & 3 months\\
{PEMS08}   & Traffic flow & 17856 & 170 &5mins & 2 months\\
{Electricity} & Electricity consumption & 2208 & 336 & 60mins & 3 months\\
\bottomrule
\end{tabular}
\vspace{-0.3cm}
\end{table*}

\textbf{Baselines}.
We select various baselines from traditional to deep learning based models that have official public codes. Historical Average (HA), VAR~\cite{VAR}, and SVR~\cite{SVR} are traditional methods. FC-LSTM~\cite{2014Seq2Seq}, DCRNN~\cite{2017DCRNN}, Graph WaveNet~\cite{GWNet},
ASTGCN~\cite{2019ASTGCN}, STSGCN~\cite{2020STSGCN}, and STGCN~\cite{stgcnyu2017spatio} are deep learning methods. GMAN~\cite{2020GMAN}, MTGNN~\cite{wu2020connecting}, GTS ~\cite{shang2021discrete}, STEP~\cite{shao2022pre}, 
STID~\cite{stidshao2022spatial} and PatchTST~\cite{patchTst} are recent state-of-the-art
works.

\textbf{Metric}. All baselines are evaluated by three
commonly used metrics in multivariate time series forecasting,
including Mean Absolute Error (MAE), Root Mean Squared Error
(RMSE) and Mean Absolute Percentage Error (MAPE).

\textbf{Implementation Details}. 
For the traffic datasets mentioned above, the input length is $L=2016$, which corresponds to the data from the past week. 
The segment length is $L_s=12$, the stride length is $l=8$, and  $P=252$.
For Electricity, the input length is $L=168$, segment length $L_s=24$, stride length $l=12$ and $P=12$.
We conducted experiments to forecast the
next 12 time steps using the same settings as the baselines. 
The dimension of the latent representations $d=96$. The \encoder uses 4 layers of Transformer blocks, and the
\agg uses 1 Transformer layer with randomly initialized target positional embedding and ranking positional embeddings. The hyper-parameters of GraphWaveNet are the same as their papers \cite{GWNet}. For the \retr, we first retrieve $K_n=5$ similar time series and then retrieve $K_s=10$ similar segments for the \agg.  
The \retr does not have parameters and is directly implemented. 
The optimizer is  Adam and the learning rate is 0.001.
\Name uses a linear layer to project $\mathbf{AO}_{P+1}$ into the forecasting dimension such as 12 for the next 12 points prediction. 
We perform significance tests (t-test with a p-value $<$ 0.05) on the experimental results. 
More detailed implementation and optimization settings could be found in Appx. \ref{appendix:opti_settings}. The code, required libraries of \Name  and the datasets 
are available in the repository \cite{ourcode}. 
\begin{table*}[h!]
\small
\renewcommand\arraystretch{0.2}
    \centering
    \caption{MTS forecasting on the PEMS-BAY,  PEMS04, PEMS07, PEMS08 and Electricity datasets. Numbers marked with $^*$ indicate that the improvement is statistically significant compared with the best baseline ~(t-test with p-value$<0.05$)}.
    \begin{tabular}{ccccr|ccr|ccr}
      \toprule
      \multirow{2}*{\textbf{Datasets}} &\multirow{2}*{\textbf{Methods}} & \multicolumn{3}{c}{\textbf{Horizon 3}} & \multicolumn{3}{c}{\textbf{Horizon 6}}& \multicolumn{3}{c}{\textbf{Horizon 12}}\\ 
      \cmidrule(r){3-5} \cmidrule(r){6-8} \cmidrule(r){9-11}
      &  & MAE & RMSE & MAPE & MAE & RMSE & MAPE & MAE & RMSE & MAPE\\
    \midrule
    \multirow{14}*{\textbf{PEMS-BAY}} 
      &HA              & 1.89  & 4.30  & 4.16\%        & 2.50  & 5.82  & 5.62\%       & 3.31  & 7.54  & 7.65\% \\ 
      &VAR             & 1.74  & 3.16  & 3.60\%        & 2.32  & 4.25  & 5.00\%       & 2.93  & 5.44  & 6.50\% \\ 
      &SVR             & 1.85  & 3.59  & 3.80\%        & 2.48  & 5.18  & 5.50\%       & 3.28  & 7.08  & 8.00\% \\ 
      &FC-LSTM         & 2.05  & 4.19  & 4.80\%        & 2.20  & 4.55  & 5.20\%       & 2.37  & 4.96  & 5.70\% \\ 
      &DCRNN           & 1.38  & 2.95  & 2.90\%        & 1.74  & 3.97  & 3.90\%       & 2.07  & 4.74  & 4.90\% \\ 
      &STGCN           & 1.36  & 2.96  & 2.90\%        & 1.81  & 4.27  & 4.17\%       & 2.49  & 5.69  & 5.79\% \\ 
      &Graph WaveNet   & 1.30  & 2.74  & 2.73\%        & 1.63  & 3.70  & 3.67\%       & 1.95  & 4.52  & 4.63\% \\
      &ASTGCN          & 1.52  & 3.13  & 3.22\%        & 2.01  & 4.27  & 4.48\%       & 2.61  & 5.42  & 6.00\% \\  
      &STSGCN          & 1.44  & 3.01  & 3.04\%        & 1.83  & 4.18  & 4.17\%       & 2.26  & 5.21  & 5.40\% \\  
      &GMAN            & 1.34  & 2.91  & 2.86\%        & 1.63  & 3.76  & 3.68\%       & 1.86  & 4.32  & 4.37\% \\  
      &MTGNN           & 1.32  & 2.79  & 2.77\%        & 1.65  & 3.74  & 3.69\%       & 1.94  & 4.49  & 4.53\% \\  
      &GTS             & 1.34  & 2.83  & 2.82\%        & 1.66  & 3.78  & 3.77\%       & 1.95  & 4.43  & 4.58\% \\
      &STID            & 1.30  & 2.81  &2.73\%         & 1.62  & 3.72  & 3.68\%       & 1.89  & 4.40  & 4.47\% \\
      &STEP            & 1.26  & 2.73  &2.59\%        & 1.55  &3.58  & 3.43\%       & 1.79  & 4.20  & 4.18\% \\   
      &PatchTST          & 1.28  & 2.77  &2.71\%         & 1.60  & 3.63  & 3.61\%       & 1.89  & 4.42  & 4.43\% \\  
    \cmidrule(r){2-11}
      &\Name           & \textbf{1.25}  & \textbf{2.68}$^*$ & \textbf{2.57}\%         & \textbf{1.53} & \textbf{3.54}    & \textbf{3.38\%}$^*$       & \textbf{1.75}$^*$ & \textbf{4.18} & \textbf{4.13\%}

      \\ 

    \midrule
    \color{black}{\multirow{14}*{\textbf{PEMS04}}}
      &HA              & 28.92  & 42.69  & 20.31\%        & 33.73  & 49.37  & 24.01\%       & 46.97  & 67.43  & 35.11\% \\ 
      &VAR             & 21.94  & 34.30  & 16.42\%        & 23.72  & 36.58  & 18.02\%        & 26.76  & 40.28  & 20.94\% \\ 
      &SVR             & 22.52  & 35.30  & 14.71\%        & 27.63  & 42.23  & 18.29\%       & 37.86  & 56.01  & 26.72\% \\ 
      &FC-LSTM         & 21.42  & 33.37  & 15.32\%        & 25.83  & 39.10  & 20.35\%       & 36.41  & 50.73  & 29.92\% \\ 
      &DCRNN           & 20.34  & 31.94  & 13.65\%        & 23.21  & 36.15  & 15.70\%       & 29.24  & 44.81  & 20.09\% \\ 
      &STGCN           & 19.35  & 30.76  & 12.81\%        & 21.85  & 34.43  & 14.13\%       & 26.97  & 41.11  & 16.84\% \\ 
      &Graph WaveNet   & 18.15  & 29.24  & 12.27\%        & 19.12  & 30.62  & 13.28\%       & 20.69  & 33.02  & 14.11\% \\
      &ASTGCN          & 20.15  & 31.43  & 14.03\%        & 22.09  & 34.34  & 15.47\%       & 26.03  & 40.02  & 19.17\% \\  
      &STSGCN          & 19.41  & 30.69  & 12.82\%        & 21.83  & 34.33  & 14.54\%       & 26.27  & 40.11  & 14.71\% \\  
      &GMAN            & 18.28  & 29.32  & 12.35\%        & 18.75  & 30.77  & 12.96\%       & 19.95     & 31.21  & 12.97\% \\  
      &MTGNN           & 18.22  & 30.13  & 12.47\%        & 19.27  & 32.21  & 13.09\%       & 20.93  & 34.49  & 14.02\% \\  
      &GTS            & 18.97  & 29.83   & 13.06\%        & 19.29  & 30.85  & 13.92\%       & 21.04  & 34.81  & 14.94\% \\

      &STID           & 17.51  & 28.48   & 12.00\%        & 18.29  & 29.86  & 12.46\%       & 19.58  & 31.79  & 13.38\% \\
      &STEP           & 17.34  & 28.44   &11.57\%         & 18.12 & 29.81    &12.00\%       & 19.27 & 31.33   & 12.78\% \\ 
      &PatchTST        & 17.42  & 28.67   & 11.96\%        &  18.23 & 29.91  & 12.34\%      & 19.41  & 31.71  & 13.09\% \\        
    \cmidrule(r){2-11}
    &\Name       & \textbf{17.09}$^*$ & \textbf{27.96}$^*$ & \textbf{11.56}\%        & \textbf{17.76}$^*$  & \textbf{29.11}$^*$  & \textbf{12.00}\%      & \textbf{18.66}$^*$  & \textbf{30.53}$^*$  & \textbf{12.67\%} \\      
        \midrule
    \color{black}{\multirow{14}*{\textbf{PEMS07}}}
      &HA              & 49.02   & 71.16   & 22.73\%     & 49.03   & 71.18   & 22.75\%   & 49.06   & 71.20   & 22.79\% \\ 
      &VAR             & 32.02   & 48.83   & 18.30\%     & 35.18   & 52.91   & 20.54\%   & 38.37   & 56.82   & 22.04\% \\ 
      &SVR             & 30.15   & 42.41   & 19.22\%     & 37.61   & 49.00   & 19.35\%   & 37.76   & 51.90   & 23.19\% \\ 
      &FC-LSTM         & 20.42   & 33.21   & 8.79\%       & 23.18   & 37.54   & 9.80\%    & 28.73   & 45.63   & 12.23\% \\ 
      &DCRNN           & 19.45   & 31.39   & 8.29\%      & 21.18   & 34.43   & 9.01\%     & 24.14   & 38.84   & 10.42\% \\ 
      &STGCN           & 20.33   & 32.73   & 8.68\%      & 21.66   & 35.35   & 9.16\%    & 24.16   & 39.48   & 10.26\% \\ 
      &Graph WaveNet   & 18.69   & 30.69   & 8.45\%       & 20.24   & 33.32   & 8.57\%     & 22.79   & 37.11   & 9.73\% \\
      &ASTGCN          & 19.31   & 31.68   & 8.18\%       & 20.70   & 34.52   & 8.66\%     & 22.74   & 37.94   & 9.71\%   \\
      &STSGCN          & 19.74   & 32.32   & 8.27\%       & 22.07   & 36.16   & 9.20\%     & 26.20   & 37.94   & 9.71\%    \\
      &GMAN            & 19.25   & 31.20   & 8.21\%       & 20.33   & 33.30   & 8.63\%     & 22.25   & 36.40   & 9.48\%    \\
      &MTGNN           & 19.23   & 31.15   & 8.55\%       & 20.83   & 33.93    & 9.30\%     & 23.60   & 38.10   & 10.10\%     \\
      &GTS             & 20.00   & 31.87   & 8.45\%       & 22.11   & 35.02    & 9.39\%     & 25.49   & 39.77   & 10.96\%       \\
      &STID            & 18.31  & 30.39  & 7.72\%         & 19.59  & 32.90  &  \textbf{8.30}\%       & 21.52  & 36.29  &  \textbf{9.15}\%      \\
      &STEP            & 18.56   & 30.33   & 8.35\%       & 19.69   & 32.55   & 8.56\%     & 21.79   & 35.30   & 9.86\% \\
      &PatchTST          & 18.60  & 30.42  &  8.79\%        & 20.23  & 33.21  &  8.53\%       & 22.30  & 36.68  & 9.59\%        \\          
    \cmidrule(r){2-11}
      &\Name          & \textbf{18.27}  & \textbf{30.29}  & \textbf{7.67}\%  & \textbf{19.42}$^*$  & \textbf{32.33}$^*$  & 
      8.31\%    & \textbf{20.39}$^*$  & \textbf{34.66}$^*$  & 9.45\% \\ 

        \midrule  
     
    \color{black}{\multirow{14}*{\textbf{PEMS08}}}
      &HA              & 23.52  & 34.96  & 14.72\%        & 27.67  & 40.89  & 17.37\%       & 39.28  & 56.74  & 25.17\% \\ 
      &VAR             & 19.52  & 29.73  & 12.54\%        & 22.25  & 33.30  & 14.23\%        & 26.17  & 38.97  & 17.32\% \\ 
      &SVR             & 17.93  & 27.69  & 10.95\%        & 22.41  & 34.53  & 13.97\%       & 32.11  & 47.03  & 20.99\% \\ 
      &FC-LSTM         & 17.38  & 26.27  & 12.63\%        & 21.22  & 31.97  & 17.32\%       & 30.69  & 43.96  & 25.72\% \\ 
      &DCRNN           & 15.64  & 25.48  & 10.04\%        & 17.88  & 27.63  & 11.38\%       & 22.51  & 34.21  & 14.17\% \\ 
      &STGCN           & 15.30  & 25.03  &  9.88\%        & 17.69  & 27.27  & 11.03\%       & 25.46  & 33.71  & 13.34\% \\ 
      &Graph WaveNet   & 14.02  & 22.76  &  8.95\%        & 15.24  & 24.22  &  9.57\%       & 16.67  & 26.77  & 10.86\% \\
      &ASTGCN          & 16.48  & 25.09  & 11.03\%        & 18.66  & 28.17  & 12.23\%       & 22.83  & 33.68  & 15.24\% \\  
      &STSGCN          & 15.45  & 24.39  & 10.22\%        & 16.93  & 26.53  & 10.84\%       & 19.50  & 30.43  & 12.27\% \\
      &GTS             & 14.50  & 22.97  &  9.23\%        & 15.77  & 25.08  & 10.09\%       & 19.02  & 28.25  & 11.74\%  \\
      &MTGNN           & 14.24  & 22.43  &  9.02\%        & 15.30  & 24.32  & 9.58\%       & 16.85  & 26.93  & 10.57\% \\  
      &GMAN            & 13.80  & 22.88  &  9.41\%        & 14.62  & 24.02  & 9.57\%       & 15.72  & 25.96  & 10.56\% \\  
      &STID            & 13.28  & 21.66  &  8.62\%        & 14.21  & \textbf{23.57}  & 9.24\%       & 15.58  & 25.89  & 10.33\%   \\
      &STEP            & 13.24   & 21.37  & 8.71\%         & 14.04  & 24.03 & 9.46\%       & 15.01  & 24.89   & \textbf{9.90}\%\\
    &PatchTST          & 13.31  & 21.65  &  8.65\%        & 14.09  & 23.93  & 9.41\%       & 15.53  & 26.36  & 10.05\%   \\   
    \cmidrule(r){2-11}
      &\Name           & \textbf{12.93}$^*$  & \textbf{21.12}$^*$  & \textbf{8.61\%}        & \textbf{13.57}$^*$  & 23.60  & \textbf{9.20\%}$^*$      & \textbf{14.48}$^*$  & \textbf{24.08}$^*$  & 10.01\% \\ 
      \midrule
    \color{black}{\multirow{14}*{\textbf{Electricity}}}
      &HA              & 92.44  & 167.00  & 70.16\%        & 92.85  & 167.05  & 70.46\%       & 92.79  & 167.21  & 70.91\% \\ 
      &VAR             & 27.69  & 56.06  & 75.53\%        & 28.19  & 57.55  & 79.94\%        & 29.34  & 60.45  & 86.62\% \\ 
      &SVR             & 23.72  & 49.19  & 70.37\%        & 27.62  & 54.05  & 78.78\%       & 27.31  & 59.47  & 73.52\% \\ 
      &FC-LSTM         & 18.57  & 48.86  & 32.88\%        & 20.68  & 48.96  & 38.21\%       & 23.79  & 56.44  & 39.98\% \\ 
      &Graph WaveNet             & 21.45  & 41.09  &  57.12\%        & 23.56  & 46.95  & 63.34\%       & 24.98  & 51.97  & 62.81\% \\
      &GTS           & 18.35  & 40.45  &  32.78\%        &21.53  &44.21   &37.91\%  & 24.62  & 49.31       & 42.57\%   \\ 
      &MTGNN           & 16.78  & 36.91  &  48.16\%        & 18.43  & 42.62  & 51.31\%       & 20.49  & 48.33  & 56.25\% \\  
      &STID            & 16.08  & 34.49  &  31.95\%        & 17.87  & 41.65  & \textbf{37.80}\%       & 19.25  & 45.77  & 40.26\%   \\
      &STEP            & 16.07  & 34.49  &  31.95\%        & 17.87  & 41.65  & 37.83\%       & 19.25  & 45.77  & 40.26\%   \\      
      &PatchTST         & 17.13  & 35.47  &  32.12\%        &18.36  &42.18   &37.98\%  & 19.69  & 45.89       & 41.30\%   \\      
    \cmidrule(r){2-11}
      &\Name           & \textbf{16.01}  & \textbf{34.27}  & \textbf{31.79\%}      & \textbf{17.56}  & \textbf{41.64} & 37.81\%      & \textbf{19.23}  & \textbf{44.52}$^*$  & \textbf{39.74}\%$^*$ \\          
      \bottomrule
    \end{tabular}
    \label{tab:main}
    \vspace{-4mm}
  \end{table*}
\vspace{-1em}

\subsection{Main Results}
As shown in Table \ref{tab:main}, the
best results are highlighted in bold.
are underlined. 
\Name consistently achieves the best performance in almost horizons in all datasets, which indicates the effectiveness of our framework.  
Traditional methods such as HA, VAR, and SVR perform worst because of their strong assumptions about the data, \eg stationary or linear. 
FC-LSTM, a classic recurrent neural network for sequential data, can not perform well since it only considers temporal features, ignoring the dependencies between MTS.
 Recently proposed spatial-temporal models overcome these shortcomings and make considerable progress.  STGCN, DCRNN, and Graph WaveNet are three typical  spatial-temporal coupling models among them.
Graph WaveNet combines GNN and Gated TCN to form a spatial-temporal layer, DCRNN replaces the fully connected layer in GRU by diffusion convolution to get a diffusion convolutional GRU.
The latest works that leverage MTS short-term history, such as ASTGCN, STSGCN, GMAN, MTGNN, GTS  have promising performance due to their refined data assumptions and reasonable model architecture. However, they can not consistently outperform other baselines.
Among them, MTGNN replaces the GNN and gated TCN in Graph WaveNet with a mix-hop propagation layer~\cite{2019MixHop} and dilated inception layer.
GMAN performs better in long-term prediction thanks to the attention mechanism's powerful ability to capture long-term dependency. 
DGCRN captures the dynamic characteristics of the spatial topology and has relatively good results. 
In the STID method, the spatial and temporal identity information is captured using MLP and has mediocre performance. 
STEP and PatchTST encode univariate long-term history into hidden states in forecasting and STEP performs better, but the useful spatial information in the long-term is still ignored. We observe that no single state-of-the-art method achieves the highest accuracy across all datasets. However, \Name consistently outperforms the aforementioned baselines in nearly all cases. Compared to typical STGNN models, LMHR achieves an average improvement of approximately 5.23\%, 8.72\%, and 12.55\% on horizons 3, 6, and 12 over all metrics, respectively, with the improvement of 10.72\% performance of the average 12 horizons. Similarly, compared to state-of-the-art models, 
LMHR demonstrates an average improvement of about 2.34\%, 3.16\%, and 4.76\% on horizons 3, 6, and 12 over all metrics, respectively, with the improvement of 4.12\% performance of the average 12 horizons.
We attribute this success to the \Name framework which could effectively leverage the long-term multivariate representations through the integration of the \retr and \agg components, supported by the powerful capabilities of the \encoder.

\subsection{Performance on Rapidly Changing Pattern Predictions}

Accurately forecasting rapidly changing patterns is critical for effective decision-making in the time series domain. For example, in traffic datasets, these patterns often correspond to peak hours, sudden events such as accidents, traffic congestion, road closures, or unexpected surges in demand, making accurate predictions essential for traffic management and planning. To evaluate \Name's performance on these patterns against state-of-the-art and baseline models, we focus on metrics specifically calculated for these scenarios in this subsection. We define rapidly changing patterns by calculating the local standard deviation over a sliding window of size \(T_f\). The local standard deviation for a given node \(i\) at time step \(t\) is computed as follows:

\begin{equation}
    \text{std}_t^i = \sqrt{\frac{1}{T_f} \sum_{w=t}^{t+T_f-1} (x^i_w - \mu^i_w)^2},
\end{equation}
where \(i \in N\) is the index of nodes, and \(\mu^i_w\) is the mean of the time series within the window:
\begin{equation}
    \mu^i_w = \frac{1}{T_f} \sum_{w=t}^{t+T_f-1} x^i_w.
\end{equation}

Then we select the top \(p\%\) of time steps with the highest \(\text{std}_t^i\) values as rapidly changing patterns for each node with high variability in the data:
\begin{equation}
    \{ t \ | \ \text{std}_t^i \geq \text{threshold} \},
\end{equation}
where the threshold corresponds to the \((100 - p)\%\)-th percentile of \(\text{std}_t^i\).

For these selected patterns at time step \(t\), we calculate the average MAE, RMSE, and MAPE over the sliding window size \(T_f\). As shown in Table \ref{tab:rapid_change}, results from the PEMS04 dataset demonstrate that \Name achieves the highest prediction accuracy compared to Graph WaveNet and STEP on the selected patterns. The average improvement of \Name over Graph WaveNet ranges from 9.80\%-10.91\% on top 10\%-30\% rapid changing patterns, which can be attributed to \Name's retrieval-based approach that leverages long-term MTS representations to enhance prediction performance. 

The percentage improvement of STEP over Graph WaveNet further illustrates that state-of-the-art models provide only marginal enhancements on these challenging patterns, with improvements of 2.97\% and 3.31\% on the top 10\% and 20\% patterns, respectively.  It also conveys that the STEP's performance gains compared with Graph WavenNet are primarily derived from the remaining stable and flat patterns. 
\Name maintains robust performance even for patterns with the top 10\% variability and fluctuations compared with STEP. 
Visualizations of predictions in Sec. \ref{sec:visulization} provide intuitive evidence to support this finding further.
\begin{table}[h]
\centering
\caption{Performance on rapidly changing patterns predictions.}
\label{tab:rapid_change}
\small
\begin{tabular}{c|c|c|c|c|c}    
\toprule
\textbf{\( p\% \)} &  \makecell{\textbf{Model} \\ \textbf{Name}} & \textbf{MAE} & \textbf{RMSE} & \textbf{MAPE } & 
\makecell{\textbf{Avg. Imp. } \\ \textbf{on Graph}\\  \textbf{WaveNet}} \\ 
\midrule
\multirow{3}{*}{10\%} & \makecell{Graph \\ WaveNet}   & 36.08 & 50.06 & 17.98\% & - \\
                      & STEP   & 35.18 & 48.63 & 17.35\% & 2.97\% \\
                      & \Name   & 32.52 & 46.04 & 15.93\% & 9.80\% \\
\midrule
\multirow{3}{*}{20\%} & \makecell{Graph \\ WaveNet}   & 32.03 & 43.72 & 15.11\% & - \\
                      & STEP   & 31.35 & 42.06 & 14.50\% & 3.31\% \\
                      & \Name   & 27.90 & 40.17 & 13.34\% & 10.91\% \\
\midrule
\multirow{3}{*}{30\%} & \makecell{Graph \\ WaveNet}    & 29.60 & 40.15 & 13.79\% & - \\
                      & STEP   & 28.07 & 38.72 & 13.23\% & 4.27\% \\
                      & \Name   & 26.60 & 36.73 & 12.30\% & 9.83\% \\
\midrule                    
\multirow{3}{*}{100\%} & \makecell{Graph \\ WaveNet}    & 19.83 & 29.05 & 13.99\% & - \\
                      & STEP   & 18.83 & 27.81 & 13.41\% & 4.49\% \\
                      & \Name   & 17.82 & 26.54 & 12.03\% & 10.92\% \\
\bottomrule
\end{tabular}
\end{table}

\subsection{Visualization}
\label{sec:visulization}
In this subsection, we inspect the performance of \Name intuitively, including the retrieved information of \retr, the learned attention weights of \agg and the final predictions of \Name. We conduct experiments on the PEMS04 test dataset and randomly select a sample from it. 

\begin{figure}
 \centering
\includegraphics[width=\linewidth]{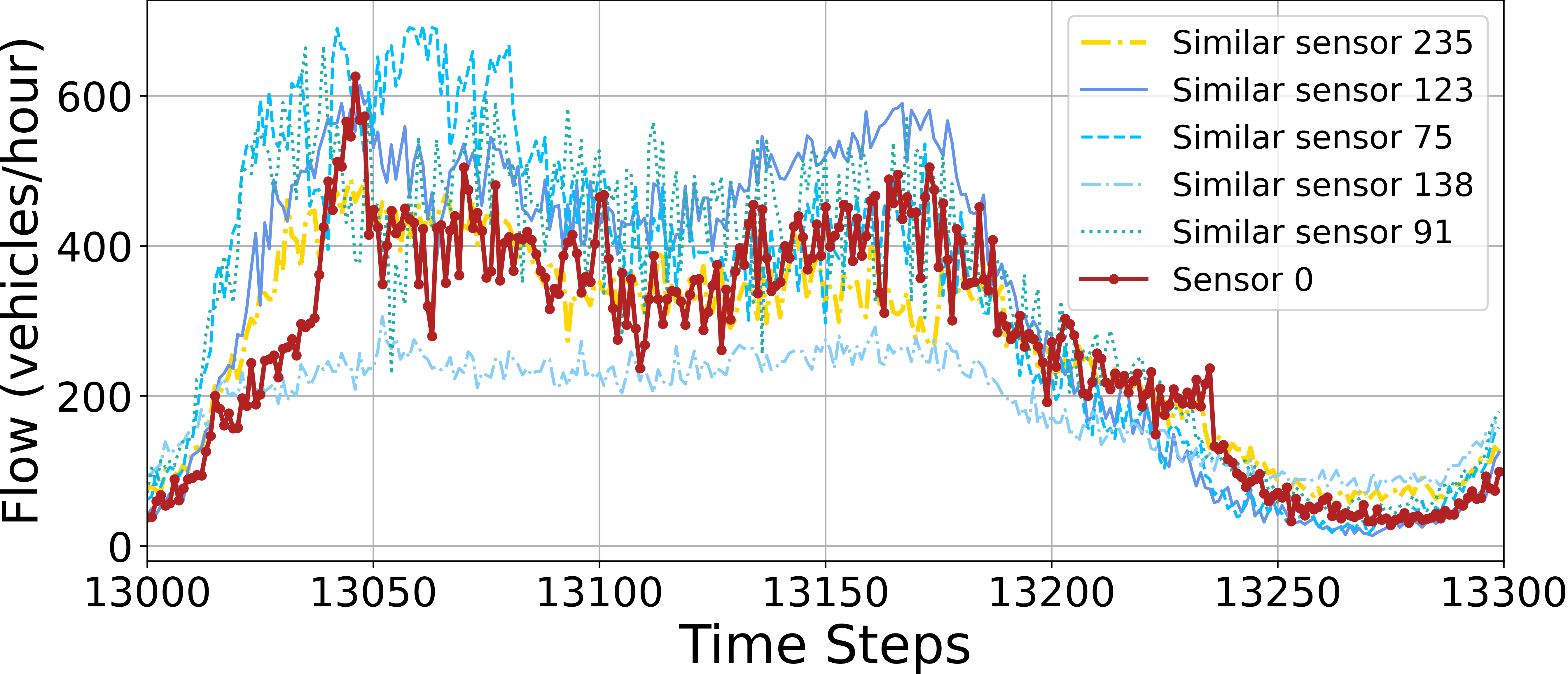}
\caption{Retrieved similar sensors from series-level of \retr. The top 5 retrieved sensors have similar long-term patterns
compared with sensor 0.} 
    \label{vis_retr_node}
 \includegraphics[width=\linewidth]{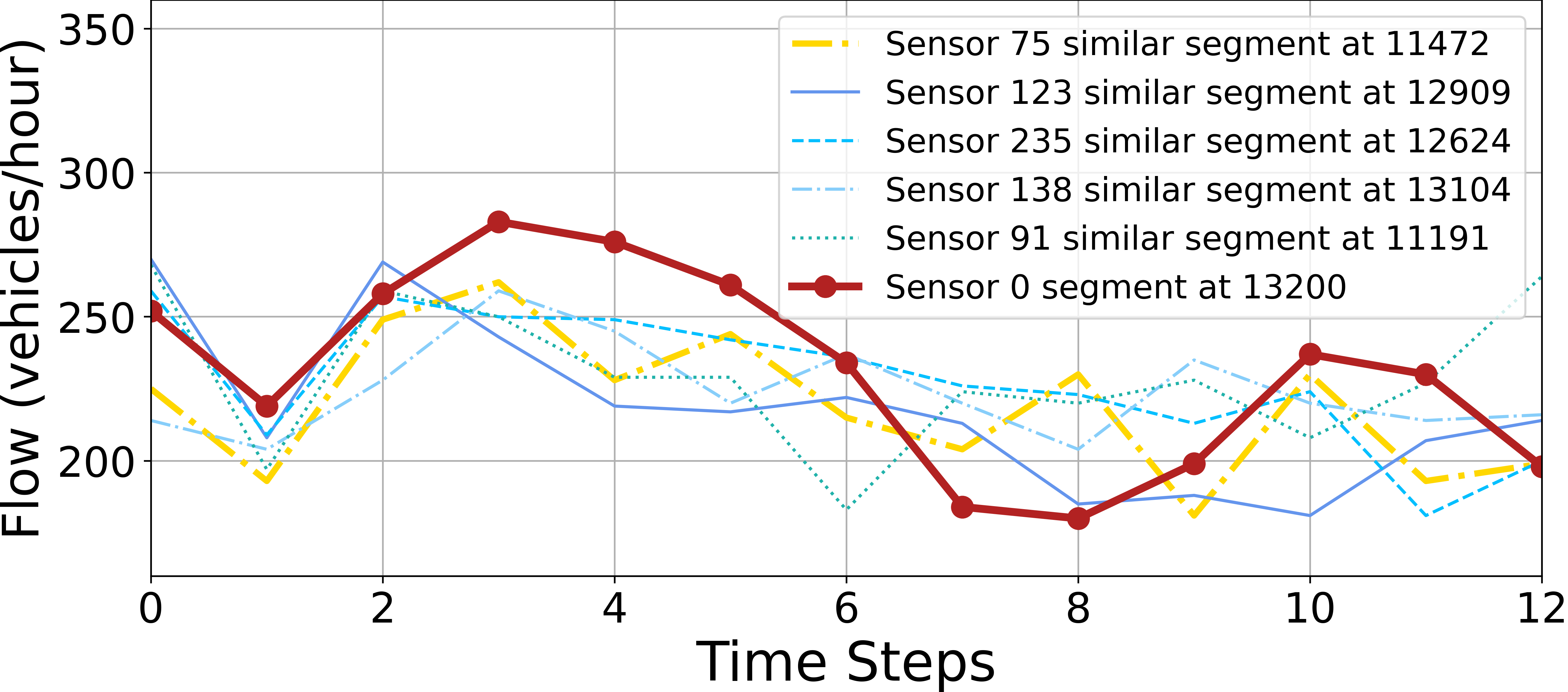}
 \caption{Retrieved  similar historical segments from segment-level of \retr. The top 5 retrieved segments have similar trends and amplitudes compared
with the target segment and are sparsely distributed over long-term MTS history. }
 \label{vis_retr_seg}
\end{figure}

\textbf{Series and Segments of the Retrieved Representations in \retr}. 
 To inspect the retrieval performance of \retr, we first plot the top 5 corresponding series of retrieved representations from the series-level. As shown in Fig. \ref{vis_retr_node}, the top 5 retrieved series  have similar long-term patterns compared with the series from sensor 0. Furthermore, it also indicates \retr
has the potential to identify the adjacent sensors from series-level. For those retrieved representations, the sensors they belong to are adjacent to sensor 0, and there would exist inherent spatial dependency between these sensors. What's more, the segments
with similar representations are more likely to be followed
by the same future series.
From the segment-level, we plot the top 5 corresponding retrieved segments from similar series with their positions in Fig. \ref{vis_retr_seg}. These segments have the same trend and amplitude compared with the target segment, and they are sparsely distributed from the long-term history.  These results show the  non-parametric \retr has the strong
ability to extract sparsely distributed and useful representations from the
long-term history, and the spatial-biased
temporal dependency is extracted explicitly. 

\begin{figure*}
\centering
 \setlength{\belowcaptionskip}{-0.5cm}
\includegraphics[width=0.7\linewidth]{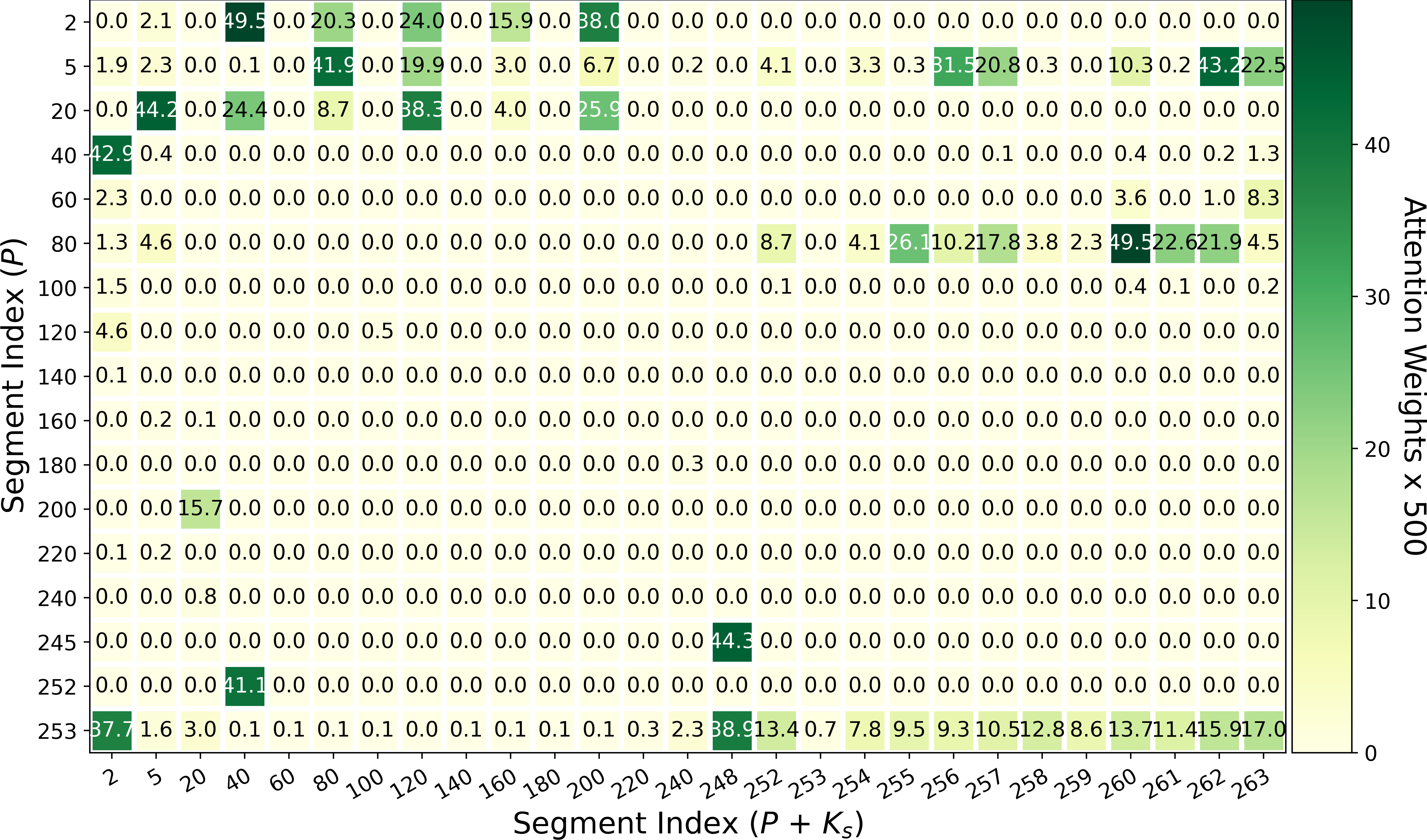}
\caption{Visualization of the attention weights demonstrates that \retr successfully retrieves useful yet sparsely distributed representations, while \agg has the capability to selectively merge them.}
\label{vis_aggr}
\end{figure*} 

\textbf{Learned Attention Weights of \agg}. In this subsection, we inspect the averaged attention weights of multiple heads in \agg to illustrate how it aggregates different valuable representations for forecasting.  
As shown in Fig. \ref{vis_aggr}, we plot the partial pairwise attention scores among \agg's inputs including 252 long-term historical segments of sensor 0, the 253rd target positional embedding, and 10 retrieved similar segments (254 to 263) from other sensors. The attention score is randomly chosen from one instance of the PEMS04 dataset. The attention score is higher if the color is deeper. 
The final representation of 253rd position from \agg, which fuses information containing long-term temporal dependencies and spatial-temporal relations, is used in final forecasting.

The first impression of Fig. \ref{vis_aggr} is that many zero scores exist, which shows the sparsity of valuable information from the long-term MTS history. When we look at the attention scores for the 253rd target position, it shows several characteristics. The 253rd position pays more attention to the 2nd, 248th, 252nd, and 254th-263rd segments than the remaining segments, such as the 40th segment. Since each segment contains 12 timestep of 1 hour, the second segment is the corresponding daily moment of one week ago (one week contains 252 segments). The attention score of 2nd position for the 253rd position indicates the underlying periodicity of sensor 0.  The 252nd and 248th segments are the recent two segments hour, which shows the influence of previous short-term history on the forecasting moment. When we take a look at the retrieved segments from other sensors, we can find that these scores are generally higher than the 40th to the 240th of sensor 0. On the one hand, these attention scores of retrieved segments show the effectiveness of the \retr. On the other hand, the value sequence of these scores is not the same as the retrieved ranking sequence, i.e. the 254th score on the 253rd row is lower than the 263rd score. This phenomenon demonstrates the necessity of \agg to selectively choose valuable information from retrieved segments representations. 

\begin{figure}
\centering
 \setlength{\abovecaptionskip}{-0.cm}
 \setlength{\belowcaptionskip}{-0.5cm}
 \includegraphics[width=1\linewidth]{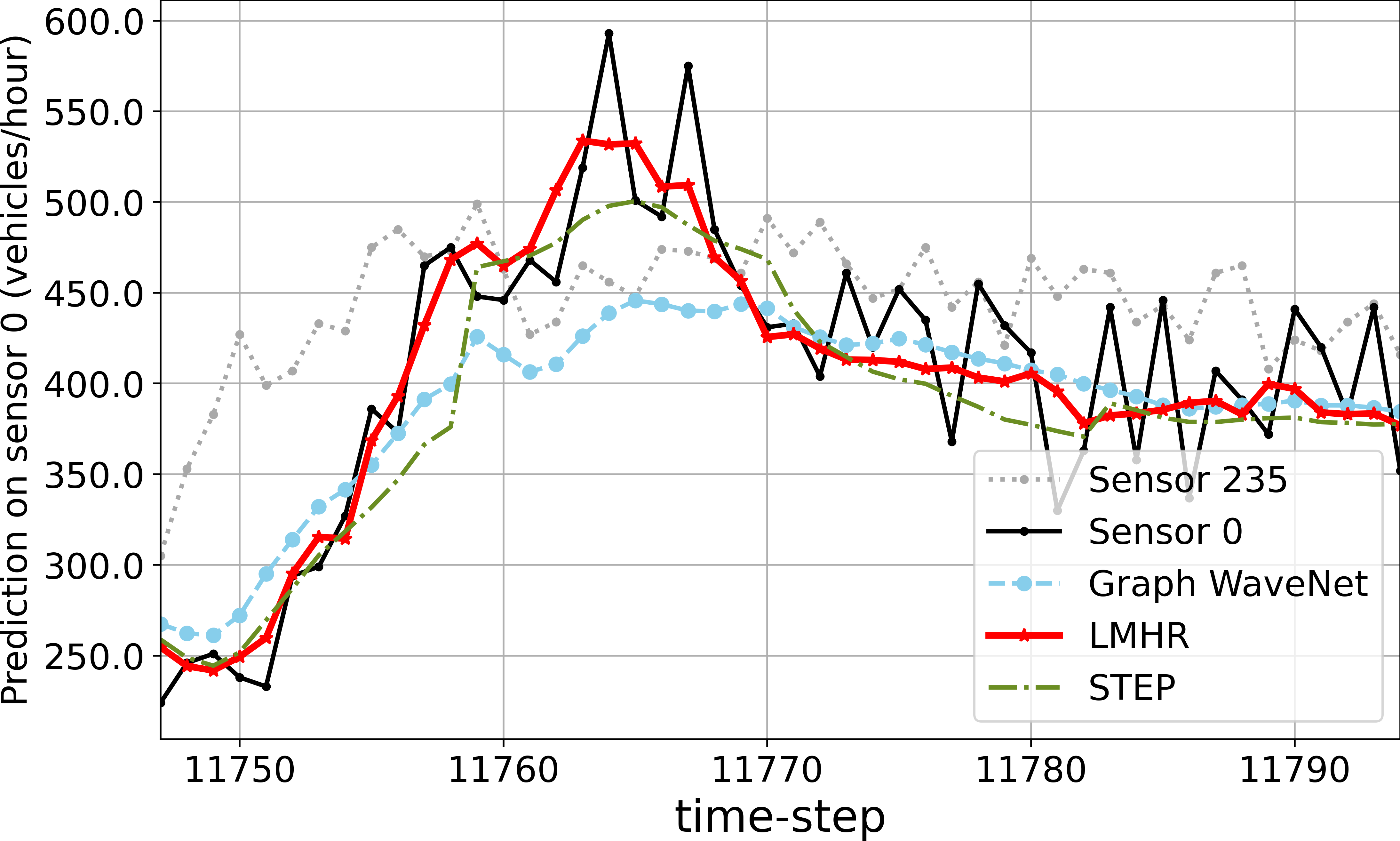}
 \caption{Predictions from \Name , Graph WaveNet and STEP  on sensor 0 show that \Name can rapidly capture the wave trend with less influence from short-term adjacent sensors.} 
\label{vis_compare}
\end{figure}

\textbf{Visualization of Predictions}.
In this subsection, we randomly select a predicted time window from sensor 0 to compare the forecasting performance of \Name, its backbone model Graph WaveNet, and STEP. As shown in Fig. \ref{vis_compare}, the black line represents the ground truth for sensor 0, while the grey line corresponds to its adjacent sensor 235. Predictions for sensor 0 from various models are also included in the figure, and it's challenging to predict sensor 0, especially before the 11773rd time step due to the complex and rapid change patterns.
We observe that \Name (red) produces predictions closer to sensor 0's ground truth (black) compared to Graph WaveNet (blue) and STEP (green), particularly in the time window around the 11755th to 11773rd steps. Interestingly, Graph WaveNet appears to align more closely with sensor 235 (grey) before the 11773rd time step, indicating the unexpected influence of short-term historical data from neighboring sensors introduced by the STGNN structure. 
In contrast, \Name quickly adapts to the wave of sensor 0 before the 11773rd step and delivers better predictions than STEP, even when the patterns change rapidly. Based on the above observations, the \Name gives more robust and accurate predictions than its backbone Graph WaveNet and STEP. With the help of long-term history and retrieved similar representations, the \Name responds quickly to the rapid changes of each sensor and relieves the short-term dependency on adjacent time series.

\subsection{Ablation Study}
In this subsection, we conduct detailed experiments to verify the impacts of different components in \Name.
{
\begin{table*}[ht]
\centering
\small
\caption{Ablation study on PEMS04.}
\begin{tabular}{cccr|ccr|ccr}
\toprule
\multirow{2}*{\textbf{Variants}}    & \multicolumn{3}{c}{\textbf{Horizon 3}}     & \multicolumn{3}{c}{\textbf{Horizon 6}}   & \multicolumn{3}{c}{\textbf{Horizon 12}}\\ 
            \cmidrule(r){2-4} \cmidrule(r){5-7} \cmidrule(r){8-10}
        & MAE & RMSE & MAPE & MAE & RMSE & MAPE & MAE & RMSE & MAPE\\
\midrule
Graph WaveNet & 18.15  & 29.24  & 12.27\%         & 19.12  & 30.62     & 13.28\%         & 20.69  & 33.02  & 14.11\% \\
\Name       & \textbf{17.09} & \textbf{27.96}  & \textbf{11.56}\%        & \textbf{17.76}  & \textbf{29.11}  & \textbf{12.00}\%      & \textbf{18.66}  & \textbf{30.53}  & \textbf{12.67\%}\\
\hline
\color{black}\textit{w/o \agg} & 17.40  & 28.37  & 11.90\%  & 17.98  & 29.32  & 12.20\%  & 18.89 & 31.73  & 12.82\% \\
\color{black}\textit{w/o $H_P^i$ representation}  & 17.31  & 28.20 & 11.75\%   & 17.83 & 29.82  & 12.05\% & 18.81  & 31.67  & 12.95\% \\
\color{black}\textit{w/o STGNN} & 18.14  & 28.84  & 12.56\%  & 18.43  & 29.98  & 12.66\% & 19.84 & 30.85  & 13.25\% \\
\color{black}\textit{w/o Graph Learning} &17.43 & 28.50 &11.79\% &18.07 & 29.68 &12.19\% &18.99 &31.68 &12.93\% \\
\textit{w/o overlap segmentation}   & 17.16  & 27.96  & 11.63\%       & 17.95  & 29.90  & 12.12\%      & 18.83  &30.97  & 12.80\% \\
 \hline
\textit{\Name with DCRNN}  & 18.03  & 28.79  & 12.48\%       & 18.76  & 30.81  & 13.20\%     &22.74  & 34.59  & 16.68\%\\ 
DCRNN & 20.34  & 31.94  & 13.65\%        & 23.21  & 36.15  & 15.70\%       & 29.24  & 44.81  & 20.09\% \\
\bottomrule
\end{tabular}
\label{tab:ablation}
\end{table*}
}

\textbf{Important Components}.
We analyze the performance of \Name by dropping or replacing one component based on the PEMS04 dataset. \textit{w/o \agg} cuts the \agg. \textit{w/o $H_P^i$ representation} means the performance of dropping the Encoder's long-term contextual representation from the node itself. \textit{w/o STGNN} is designed to directly perform forecasting by replacing STGNN with a linear layer connected with the output of \agg. \textit{w/o Graph Learning} drops the Graph Structure Learning module and directly uses the adjacency matrix from the \retr as the graph structure for the STGNN. 
\textit{w/o} overlap segmentation means the hard break of segmentation with the $l=12$.
\textit{\Name with DCRNN} replaces the backbone Graph WaveNet with DCRNN.

As shown in Table \ref{tab:ablation}, the average performance experienced a 3.8\% decline by using \textit{w/o \agg}, which shows the necessity to selectively choose
valuable information from retrieved segment representations. 
When we drop the last segment-level representation $\mathbf{H}_P^i$  from the \encoder, the performance decreases 2.4\% indicating the need for long-term contextual representation, which contains the periodicity from the node itself. \textit{w/o STGNN} performs worse than \Name but still gives reasonable results, showing that the fused representation is meaningful and captures the trend and characteristics of each time series. \textit{w/o Graph Learning} predicts less accurately showing the dynamically generated graph structure captures underlying dependency among MTS. 
\textit{w/o} overlap segmentation performances slightly worse than \Name, the reason is that the soft break keeps the useful temporal patterns to the fullest extent with higher resolution.
\textit{\Name with DCRNN} exhibits significant improvement compared with DCRNN showing the generality of the proposed \Name framework. 

\begin{figure}
\centering
 \setlength{\belowcaptionskip}{-0.3cm}
\includegraphics[width=1\linewidth]{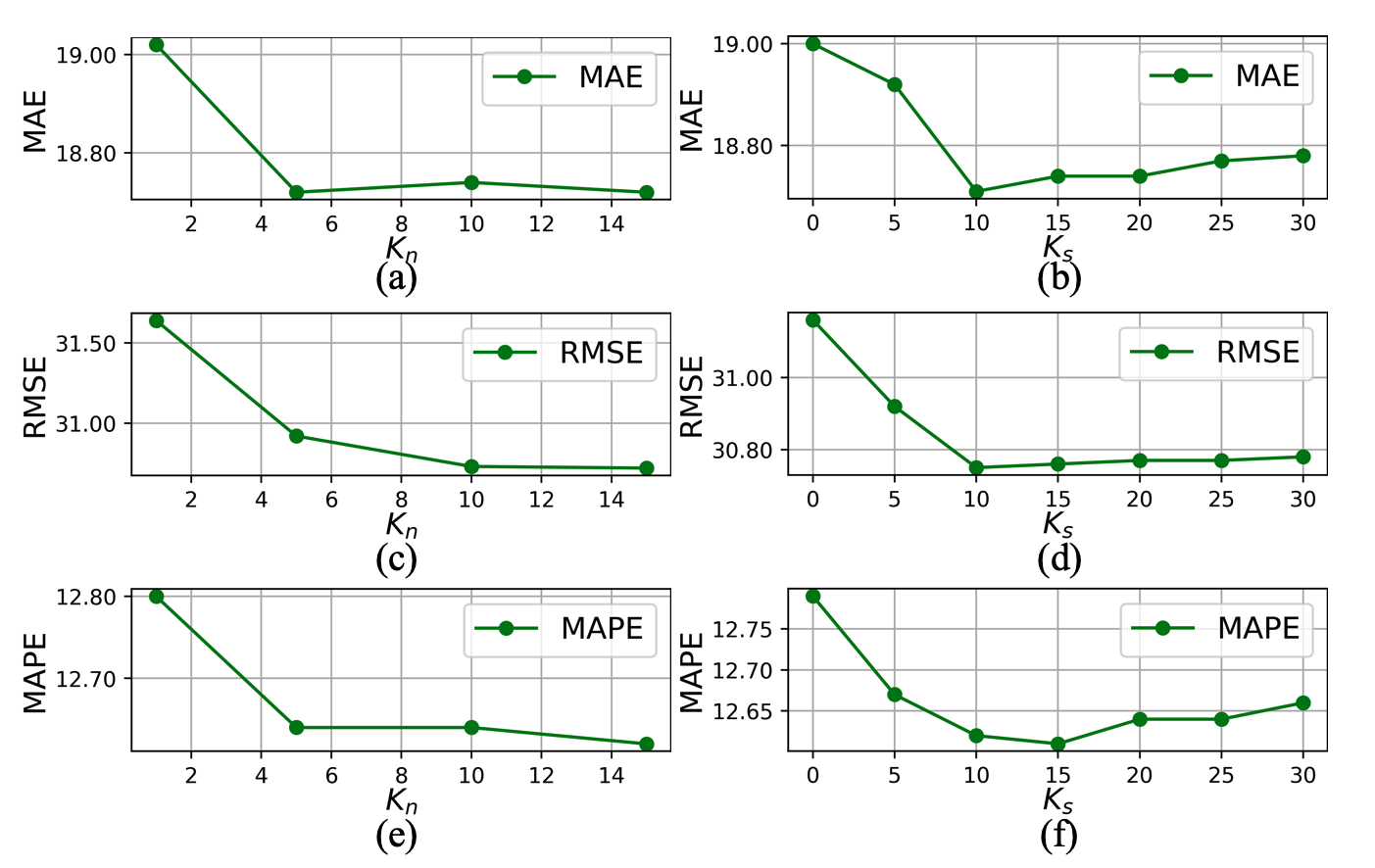}
\caption{The impact of the number of retrieved similar series \(K_n\) and segment representations \(K_s\) at horizon 12.
(a) MAE with $K_n$. (b) MAE with $K_s$. (c) RMSE with $K_n$. (d) RMSE with $K_s$. (e) MAPE with $K_n$. (f) MAPE with $K_s$. These visualizations demonstrate that \Name remains stable when \(K_n\) and \(K_s\) exceed 5 and 10, respectively.}
\label{fig:kn ks hyper}
\end{figure}

\textbf{Hyper-parameter Study}.
In this subsection, we conduct experiments on PEMS04 dataset to analyze the impacts of two key Hyper-parameters in \Name which are the number of retrieved similar series and segments as $K_n$ and $K_s$. As shown in Fig. \ref{fig:kn ks hyper} (a), (c) and (e), the \Name achieves better overall forecasting accuracy with larger $K_n$ which shows the potentially valuable information from adjacent sensors. In Fig. \ref{fig:kn ks hyper} (b), (d) and (f), the evaluation metrics become better when the $K_s$ increases. It verifies the intuition of mining useful segment representations from long-term MTS. The evaluation metrics are slightly worse when $K_s$ is larger than 15 because of introducing more trivial information.  
\vspace{-0.2cm}
\subsection{Efficiency Study}
In this section, we evaluate the efficiency of \Name, its backbone architecture Graph WaveNet, and the state-of-the-art method STEP in terms of average training speed, convergence epochs, and inference speed. All experiments were conducted on a computing server equipped with an RTX6000 GPU, and the results are presented in Table \ref{tab:efficiency} using the PEMS04 dataset.

The original Graph WaveNet, configured with a 12-timestep input, requires 103.3 seconds per epoch for training. To extend its capability, we modified Graph WaveNet to accept a 2016-timestep input, matching the input length of STEP. This modification involves 12 layers with a kernel size of 2 in each layer, enabling coverage over the 2016-timestep horizon. However, this configuration failed to yield performance improvements over the original model and significantly increased training time to over 970 seconds per epoch, showing the growing computational complexity of traditional STGNNs with long-term input.
In contrast, the state-of-the-art STEP model requires 273.2 seconds per epoch for training using univariate historical data. Compared to STEP, \Name with the same 2016-timestep input achieves a comparable training time of 279.6 seconds per epoch. The reason is that the core component of \Name's \retr introduces no additional training requirements.
%
Moreover, leveraging the retrieved useful representations, \Name requires fewer training epochs to converge than STEP. Additionally, the \agg module introduces only 0.12MB of trainable parameters compared with STEP, enabling efficient deployment on GPUs.

In inference stage, Graph WaveNet-12 achieves an average speed of 285 timesteps per second per sensor on the entire test set (307 sensors with 10 segments retrieved per sensor). STEP achieves approximately 128 timesteps per second per sensor, whereas \Name achieves the speed of 123 timesteps per second per sensor. Although \Name is slightly slower than STEP, this speed is sufficient for most real-world applications. 
\Name remains a practical model due to its data efficiency and substantial performance improvements of over 10\% compared to Graph WaveNet and 4\% compared to STEP. Furthermore, this suggests potential for further efficiency gains if a pre-trained \encoder is utilized as STEP, as the \agg module is the only component requiring training.

\begin{table}[htb]
\centering
\caption{Training speed, convergence epochs, and inference speed comparison.}
\label{tab:efficiency}
\small
\begin{tabular}{c|c|c|c}
\toprule
\textbf{Model Name} & \makecell{\textbf{Train. Speed} \\ \textbf{(s/epoch)}} & \makecell{\textbf{Convg. } \\ \textbf{Epochs}} & \makecell{\textbf{Inf. Speed} \\ \textbf{(timesteps/s)}} \\
\midrule
Graph WaveNet-12     & 103.3 & 125 & 285 \\
Graph WaveNet-2016   & 970.8  & 130   & 165  \\
STEP                 & 273.2 & 75  & 128 \\
\Name                & 279.6 & 55  & 123 \\
\bottomrule
\end{tabular}
\end{table}

\section{Related Work}
Multivariate time series (MTS) forecasting is considered as a challenging task due to the complex interdependencies between time steps and series. Classic methods such as ARIMA, VAR and SVR can model the temporal information for prediction \cite{MTS_survey}. The prediction accuracy of these methods is limited because they only capture the linear relationships in the data. However, with advancements in technology, deep learning methods like recurrent neural network (RNN)-based models have garnered lots of attention due to their ability to capture non-linear temporal dependencies \cite{2017DCRNN}. Yet, these models still fall short in modeling spatial dependencies.
The Graph Neural Network (GNN) \cite{2017GCN} has been introduced as a novel deep learning paradigm to learn from non-Euclidean data using graph analysis methods. The structure of MTS network can naturally be modeled as a graph, where nodes in the  network are interconnected and the spatial dependencies modeling could be considered.

Specifically, STGNN is an advanced neural network model designed to capture both spatial and temporal dependencies within graph-structured data.
There are two trends for STGNN methods. The first is the graph convolutional recurrent neural network. 
DCRNN~\cite{2017DCRNN}, STMetaNet~\cite{2019STMetaNet}, and AGCRN~\cite{2020AdaptiveGCRN}
combine diffusion convolution networks and recurrent neural networks~\cite{2014GRU} with their variants. 
They follow the seq2seq framework~\cite{2014Seq2Seq}  and adopt an encoder-decoder architecture to forecast step by step. In particular, the main limitations of these RNN-based networks are computational and memory bottlenecks, which make it difficult to scale up with large networks.
The second trend is the fully spatial-temporal graph convolutional network which aims to address the computational problem caused by using recurrent units. Graph WaveNet~\cite{GWNet}, MTGNN~\cite{wu2020connecting} and STGCN~\cite{stgcnyu2017spatio}
combine graph convolutional networks and gated temporal convolutional networks with their variants. 
These methods are based on convolution operation, which facilitates parallel computation.
Moreover, the attention mechanism is widely used in many methods for modeling spatial temporal correlations, such as GMAN~\cite{2020GMAN} and ASTGCN~\cite{2019ASTGCN}. 
Although STGNNs have made significant progress, the complexity of the above STGNNs is still high because it needs to deal with both temporal and spatial dependency at every timestep. Hence, STGNNs are limited to taking short-term multivariate history as input, such as the classic settings of 12 time steps \cite{Onesize_INNLS,knnmts, JointSpatiotemporal_INNLS}. 

Recently, due to the powerful representation ability, Transformer \cite{transformervaswani2017attention} obtains the outstanding performance in the machine translation tasks and MTS forecasting tasks with the temporal dependencies modeling. \cite{shao2022pre,patchTst} find that long-term historical data of each time series could also help the final forecasting given the longer look-back window using the segmentation. And they only utilize the univariate time series to capture the long-term temporal dependencies. 
However, in MTS forecasting, the spatial information and relations over the long-term multivariate history are also important, since similar patterns could be observed and temporal dependencies would also be deeply influenced by the spatial factors in the long run. In summary, there is still no existing works to forecast MTS by modeling the spatial-temporal correlations  over the long-term multivariate history.

\section{Conclusion}
In this paper, we propose a simple yet effective \Name framework for MTS forecasting which strengthens STGNNs with spatial-temporal correlations over long-term MTS representations.
\Name hierarchically retrieves similar segment representations from series-level and segment-level, and is capable of extracting useful and sparsely distributed information from long-term history without further training. Furthermore, \Name fuses the retrieved long-term representations through \agg and short-term spatial-temporal dependencies, having a wide
respective field for better prediction. \Name also leverages the
segment-level representation to generate an adjacency matrix 
to guide the graph structure learning in case the pre-defined graph is not complete, and is a general framework that can absorb and enhance almost arbitrary STGNNs. 
We conduct extensive evaluations of \Name on five real-world datasets, demonstrating the superiority of the \Name framework and its interpretability. Experimental results show that our proposed framework outperforms typical STGNNs by 10.72\% on
the average prediction horizons and state-of-the-art methods by 4.12\%. Moreover, it improves prediction accuracy by 9.8\% on the top 10\% of rapidly changing patterns across the datasets. Future work could focus on enhancing the robustness of this framework, particularly in handling missing data, and further improving its efficiency by incorporating more informative domain knowledge.

\bibliographystyle{IEEEtran}
\normalem
\bibliography{ref}

\appendix
\subsection{In-batch computation of \retr }
\label{appendix_algo}
The in-batch retriever computation of \retr is described in Algorithm \ref{algo:hsr_new}. In this algorithm, the variables $B$ and $N$ represent the batch size and the number of series, respectively. The parameters $K_n$ and $K_s$ are the number of retrieved series-level and segments-level representations. 
The input of \retr includes the batch output $\mathcal{H} \in \mathbb{R}^{B \times N \times P \times d}$ from the \encoder, along with $K_n$ and $K_s$. The outputs of \retr are batch-level series adjacency matrix 
 $batch\_series \in \mathbb{R}^{B \times N \times N}$, and the batch-level retrieved segment representations $batch\_segs \in \mathbb{R}^{B \times N \times K_s \times d}$.

\begin{algorithm}[t!]
\small
\caption{In-batch Computation of HRetriever}
\label{algo:hsr_new}

\KwIn{
 \\
     $\mathcal{H}$: Input tensor of shape $\mathbb{R}^{B \times N \times P \times d}$\;
     $K_n$: Number of similar series to retrieve\;
     $K_s$: Number of similar segments to retrieve\;
}
\KwOut{
\\
    $batch\_series$: Batch-level series adjacency matrix of shape $\mathbb{R}^{B \times N \times N}$ \;
     $batch\_segs$: Batch-level retrieved segment representations of shape $\mathbb{R}^{B \times N \times K_s \times d}$\;
}
\vspace{0.3cm}
\SetAlgoLined 
Reshape the input tensor $\mathcal{H} \in \mathbb{R}^{B \times N \times P \times d}$ into 
$h\_fla \in \mathbb{R}^{B \times N \times (P \cdot d)}$\;

Compute the batch-level series similarity matrix 
$series\_matrix \in \mathbb{R}^{B \times N \times N}$ using 
$\text{BatchCosineSimilarity}(h\_fla, h\_fla)$\;

Identify the top $K_n$ similar series in $series\_matrix$, and store them in 
$batch\_series \in \mathbb{R}^{B \times N \times N}$\;

Extract and reshape the last segment representation for each series from $\mathcal{H}$ to a query tensor 
$segs\_query \in \mathbb{R}^{B \times N \times d}$\;

Perform a batch matrix-matrix product between $(batch\_series, h\_fla)$ and reshape the result to obtain the retrieved similar series representation 
$series\_flat \in \mathbb{R}^{B \times N \times (K_n \cdot P) \times d}$\;

Compute the batch segment-level similarity matrix 
$segs\_matrix \in \mathbb{R}^{B \times N \times K_n \times P}$ using 
$\text{BatchCosineSimilarity}(segs\_query, series\_flat)$\;

Identify and retrieve the top $K_s$ segment representations from $segs\_matrix$ and $h\_flat$ as 
$batch\_segs \in \mathbb{R}^{B \times N \times K_s \times d}$\;

Return $batch\_series$  and $batch\_segs$.

\vspace{0.3cm}
\SetKwFunction{FBS}{BatchCosineSimilarity}
\SetKwProg{Fn}{Function}{:}{}
\Fn{\FBS{$x, y$}}{
    Normalize $x$ and $y$ using their L2 norms\;
    Compute the cosine similarity between the normalized vectors $x$ and $y$ according to Equ. \ref{eq:batch_cos_similrity}\;
    \KwRet{The pairwise cosine similarity between $x$ and $y$}\;
}

\end{algorithm}

\subsection{Implementation and Optimization Settings}
\label{appendix:opti_settings}
In this subsection, we provide detailed implementation and optimization settings for \Name. All experiments were conducted on a computing server equipped with an RTX6000 GPU with 24GB of memory, using Python 3.8 and CUDA 11.3. The model was built using the commonly used time series framework \href{https://github.com/GestaltCogTeam/BasicTS}{BasicTS+} \cite{basicts}.
The code of \Name and raw data are available in the repository \cite{ourcode}.

For the \encoder, we used the PyTorch official implementation of \textit{torch.nn.Transformer} to construct the Transformer blocks. Specifically, we utilized 4 layers of Transformer blocks with an embedding dimension of $d=96$, 4 attention heads, and a dropout ratio of 0.1. The positional embeddings were initialized using a uniform distribution, while the last token was initialized using a truncated normal distribution with $\mu=0$ and $\sigma=0.02$, following the same approach as STEP \cite{shao2022pre}.

The \retr\ module is parameter-free and directly implemented, with $K_n=5$ and $K_s=10$. The \agg\ module consists of a single Transformer layer, with randomly initialized target and ranking positional embeddings, configured identically to the \encoder.  In graph structure learning module, a convolutional neural network (CNN) with two \texttt{Conv1D} layers extracts global features from raw node data. Batch normalization layers are also used to stabilize training, and a fully connected layer (\texttt{fc}) reduces the feature dimension to 96.
Another fully connected layer (\texttt{fc\_out}) aggregates concatenated node embeddings into final graph embeddings. The activation function is the ReLU function.
This module integrates dynamic graph embeddings into spatiotemporal models such as Graph WaveNet, providing flexibility and adaptability for various datasets. For the forecasting MLP layer, we used a 3-layer fully connected network with an input dimension of 96, a hidden dimension of 256, and an output dimension of 12, with ReLu activation function. The optimizer used in this paper is Adam, with a learning rate of 0.001, a weight decay of $1.0 \times 10^{-5}$, and an epsilon of $1.0 \times 10^{-8}$. 

In the traffic datasets, the input length is $L=2016$, corresponding to data from the past week, and $C=3$, which includes features derived from raw traffic data, as well as time-of-day and day-of-week indices. The segment length is $L_s=12$, the stride length is $l=8$, and $P=252$. The batch size is set as 8.
In the Electricity dataset, the input length is $L=168$, the segment length is $L_s=24$, the stride length is $l=12$, and $P=12$. Similarly, $C=3$, which includes features from raw electricity data, time-of-day, and day-of-week indices. The batch size is set as 16.

\begin{IEEEbiography}[{\includegraphics[width=1in,height=1.25in,clip,keepaspectratio]{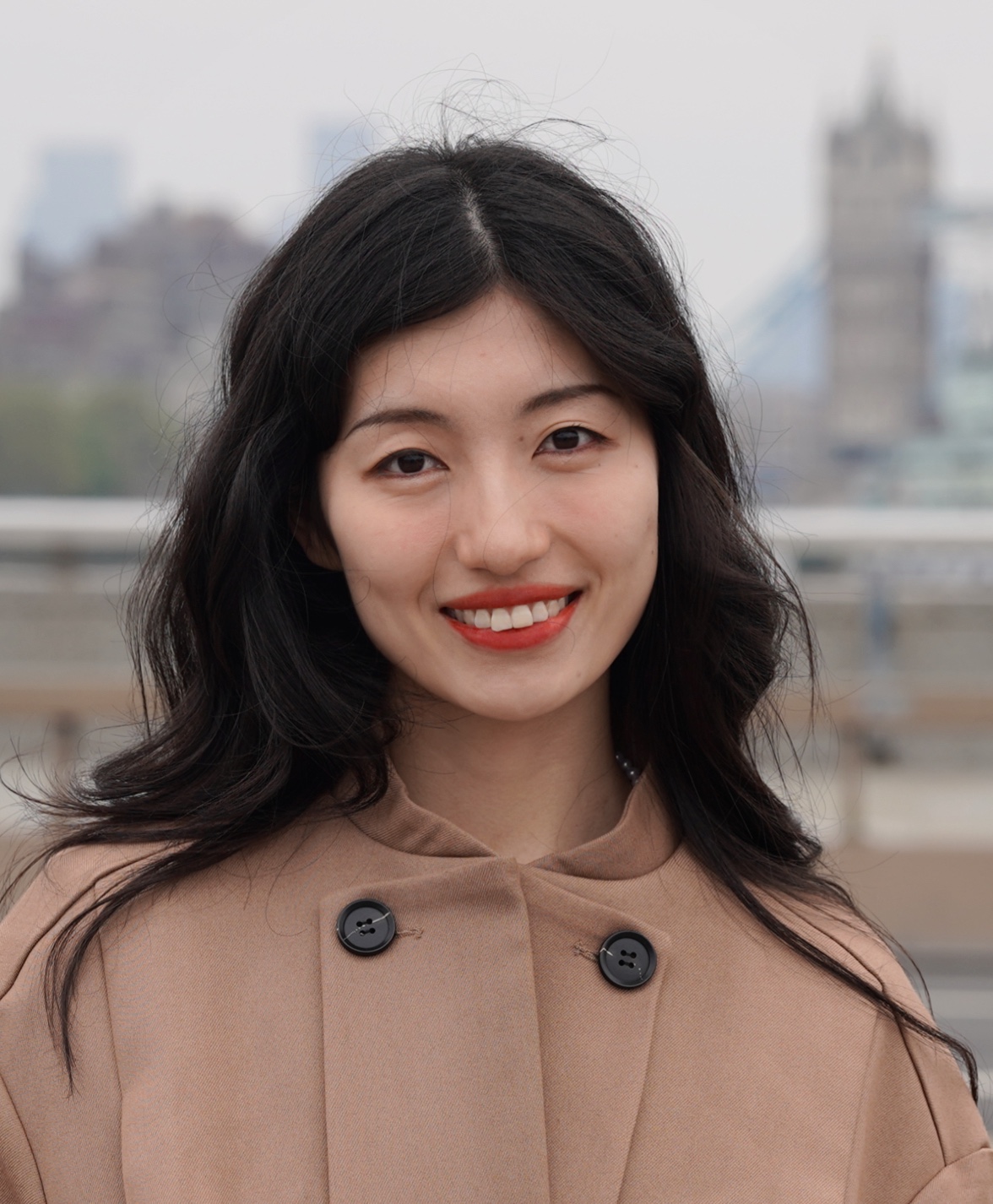}}]{Huiliang Zhang} is currently a Ph.D. candidate in the Department of Electrical and Computer Engineering at McGill University. She received her M.E. degree from Peking University, Beijing, China, in 2020 and her B.E. degree from Xidian University, Xi'an, China, in 2017.
She is interested in time series analysis, spatiotemporal data and graph modelling, and other machine learning and statistics solutions for research and applications in smart grids and intelligent transportation systems.
\end{IEEEbiography}
\begin{IEEEbiography}[{\includegraphics[width=1in,height=1.25in,clip,keepaspectratio]{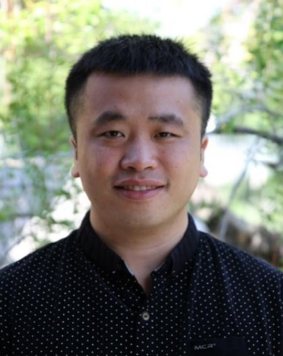}}]{Di Wu}
is currently a staff research scientist at Samsung AI center Montreal and an Adjunct Professor at McGill University. Before joining Samsung, he did postdoctoral research at Montreal MILA and Stanford University. He received the Ph.D. degree from McGill University, Montreal, Canada, in 2018 and the MSc degree from Peking University, Beijing, China, in 2013. 
Di’s research interests mainly lie in designing algorithms (e.g., Reinforcement learning, operation
research) for sequential decision-making problems and data-efficient machine learning algorithms (e.g., Transfer Learning, Meta-Learning, and Multitask Learning). He is also interested in leveraging such algorithms for applications in real-world systems (e.g., Communication Systems, Smart Grid, and Intelligent Transportation Systems)
\end{IEEEbiography}
\begin{IEEEbiography}[{\includegraphics[width=1in,height=1.25in,clip,keepaspectratio]{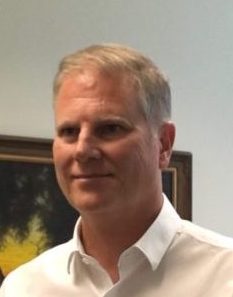}}]{Benoit Boulet} (S’88–M’92–SM’07), P.Eng., Ph.D., is Professor in the Department of Electrical and Computer Engineering at McGill University which he joined in 1998, and Director of the McGill Engine, a Technological Innovation and Entrepreneurship Centre. He is Associate Vice-Principal of McGill Innovation and Partnerships and was Associate Dean (Research \& Innovation) of the Faculty of Engineering from 2014 to 2020. Professor Boulet obtained a Bachelor's degree in applied sciences from Universit\'{e} Laval in 1990, a Master of Engineering degree from McGill University in 1992, and a Ph.D. degree from the University of Toronto in 1996, all in electrical engineering. He is a former Director and current member of the McGill Centre for Intelligent Machines where he heads the Intelligent Automation Laboratory. His research areas include the design and data-driven control of electric vehicles and renewable energy systems, machine learning applied to biomedical systems, and robust industrial control.
\end{IEEEbiography}
\begin{IEEEbiography}[{\includegraphics[width=1in,height=1.25in,clip,keepaspectratio]{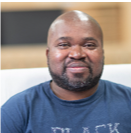}}]{Arnaud Zinflou} is currently principal research scientist at Hydro-Québec research institute and leads projects in many areas of machine learning such as computer vision, time series forecasting or representation learning. He obtained a Ph.D. from Université du Québec à Chicoutimi, Québec, Canada in 2008 and a MSc degree from Université du Québec à Montréal in 2004. He became a member of IEEE in 2008 and a senior member in 2015. He as an extensive history working with AI projects in the fields of industrial manufacturing, logistics, finance, and online retail. He is also the author and co-author of more than 50 papers, 7 book chapters and 3 patents.
\end{IEEEbiography}
\begin{IEEEbiography}[{\includegraphics[width=1in,height=1.25in,clip,keepaspectratio]{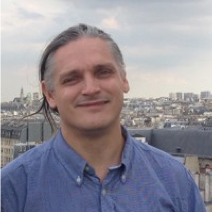}}]{Stéphane Dellacherie}  has been an engineer at Hydro-Québec since 2016 and an associate professor at Université du Québec à Montréal since 2022. Before joining Hydro-Québec, he was a visiting researcher in applied mathematics at Polytechnique Montréal (Québec) (2015-2016), a researcher in applied mathematics and scientific computing at the French Atomic Energy Commission (CEA, France) (1995-2016), and an associate researcher at the Jacques-Louis Lions Laboratory, Sorbonne University, France (2011-2015).
\end{IEEEbiography}
\begin{IEEEbiography}[{\includegraphics[width=1in,height=1.25in,clip,keepaspectratio]{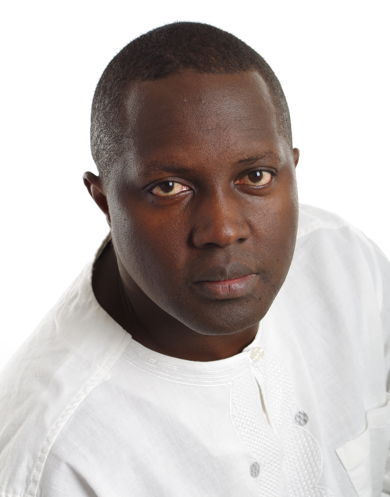}}]{Mouhamadou Makhtar Dione} received the M.Eng. degree in electrical engineering from ENSIEG/INPG, Grenoble, France, in 2000. He has been with Hydro-Québec’s Research Institute since 2011, where he is a Researcher in transmission and distribution planning in the context of high distributed energy resources integration.
\end{IEEEbiography}

\end{document}